\documentclass[acmtog]{acmart}

\usepackage{xspace}

\newcommand{\methodname}{CubePart\xspace}

\usepackage{amsfonts,bm}

\def\x{{x}}

\def\xi{{\x_i}}

\newcommand{\reffig}[1]{Figure~\ref{fig:#1}}
\newcommand{\refsec}[1]{Section~\ref{sec:#1}}

\newcommand{\lblfig}[1]{\label{fig:#1}}
\newcommand{\lblsec}[1]{\label{sec:#1}}

\newcommand{\lbltbl}[1]{\label{tbl:#1}}

\newcommand{\ignorethis}[1]{}

\def\eqref#1{equation~\ref{#1}}

\def\1{\bm{1}}

\DeclareMathAlphabet{\mathsfit}{\encodingdefault}{\sfdefault}{m}{sl}
\SetMathAlphabet{\mathsfit}{bold}{\encodingdefault}{\sfdefault}{bx}{n}

\newcommand{\ignore}[1]{}

\makeatletter
\DeclareRobustCommand\onedot{\futurelet\@let@token\@onedot}
\def\@onedot{\ifx\@let@token.\else.\null\fi\xspace}
\def\eg{e.g\onedot,\xspace}

\makeatother

\definecolor{MyDarkBlue}{rgb}{0,0.08,1}
\definecolor{MyDarkGreen}{rgb}{0.02,0.6,0.02}
\definecolor{MyDarkRed}{rgb}{0.8,0.02,0.02}
\definecolor{MyDarkOrange}{rgb}{0.40,0.2,0.02}
\definecolor{MyPurple}{RGB}{111,0,255}
\definecolor{MyRed}{rgb}{1.0,0.0,0.0}
\definecolor{MyGold}{rgb}{0.75,0.6,0.12}
\definecolor{MyDarkgray}{rgb}{0.66, 0.66, 0.66}
\definecolor{myorange}{RGB}{255,69,0}
\definecolor{derekblue}{RGB}{144,210,236}
\definecolor{newpurple}{RGB}{238, 130, 238}

\definecolor{shadecolor}{gray}{0.92}
\usepackage{listings}
\usepackage{framed}

\newif\ifhighlight
\highlightfalse

\newcommand{\revised}[1]{%
    \ifhighlight
        \textcolor{blue}{#1}%
    \else
        #1%
    \fi
}

\usepackage[ruled,lined]{algorithm2e} %

\SetAlFnt{\small}
\SetAlCapFnt{\small}
\SetAlCapNameFnt{\small}
\SetAlCapHSkip{0pt}

\usepackage{pifont}
\usepackage{makecell}
\newcommand{\cmark}{\ding{51}}
\newcommand{\xmark}{\ding{55}}

\usepackage{multirow}

\acmJournal{TOG}

\copyrightyear{2026}
\acmYear{2026}
\setcopyright{cc}
\setcctype{by}
\acmConference[SIGGRAPH Conference Papers '26]{Special Interest Group on Computer Graphics and Interactive Techniques Conference Conference Papers}{July 19--23, 2026}{Los Angeles, CA, USA}
\acmBooktitle{Special Interest Group on Computer Graphics and Interactive Techniques Conference Conference Papers (SIGGRAPH Conference Papers '26), July 19--23, 2026, Los Angeles, CA, USA}
\acmDOI{10.1145/3799902.3811117}
\acmISBN{979-8-4007-2554-8/2026/07}

\begin{document}

\title{\methodname: An Open-Vocabulary Part-Controllable 3D Generator}

\author{Yiheng Zhu}
\authornote{The first four authors contributed equally to this research.}
\email{yzhu@roblox.com}
\affiliation{
  \institution{Roblox}
  \country{USA}
}

\author{Kangle Deng}
\authornotemark[1]
\email{kdeng@roblox.com}
\affiliation{
  \institution{Roblox}
  \country{USA}
}

\author{Jean-Philippe Fauconnier}
\authornotemark[1]
\email{jfauconnier@roblox.com}
\affiliation{
  \institution{Roblox}
  \country{USA}
}

\author{Inaki Navarro}
\authornotemark[1]
\email{inavarro@roblox.com}
\affiliation{
  \institution{Roblox}
  \country{USA}
}

\author{Daiqing Li}
\email{daiqingli@roblox.com}
\affiliation{
  \institution{Roblox}
  \country{USA}
}

\author{Ava Pun}
\email{apun@andrew.cmu.edu}
\affiliation{
  \institution{Roblox}
  \country{USA}
}
\affiliation{
  \institution{Carnegie Mellon University}
  \country{USA}
}

\author{Yinan Zhang}
\email{yinan.zhang@roblox.com}
\affiliation{
  \institution{Roblox}
  \country{USA}
}

\author{Peiye Zhuang}
\email{pzhuang@roblox.com}
\affiliation{
  \institution{Roblox}
  \country{USA}
}

\author{Xiaoxia Sun}
\email{xiaoxiasun@roblox.com}
\affiliation{
  \institution{Roblox}
  \country{USA}
}

\author{Maneesh Agrawala}
\email{maneesh@cs.stanford.edu}
\affiliation{
  \institution{Roblox}
  \country{USA}
}
\affiliation{
  \institution{Stanford University}
  \country{USA}
}

\author{Kiran Bhat}
\email{kbhat@roblox.com}
\affiliation{
  \institution{Roblox}
  \country{USA}
}

\author{Tinghui Zhou}
\email{tzhou@roblox.com}
\affiliation{
  \institution{Roblox}
  \country{USA}
}

\begin{abstract}
Interactive 3D assets used in games and simulation are typically decomposed into specific semantic parts to support animation, physics, and scripted behaviors, yet most generative 3D models produce either monolithic meshes or arbitrary part decompositions that cannot be aligned with application-specific requirements. We present \methodname, a generative framework for open-vocabulary, part-controllable 3D mesh generation that exposes part structure as an explicit inference-time control signal. Given a global text prompt and a user-defined parts schema expressed as an open-ended list of part names, our method generates a set of meshes—one per schema element—that assemble into a coherent object while respecting the specified semantic structure. To enable this capability, we introduce a scalable data pipeline to construct a large open-vocabulary, part-labeled 3D dataset, along with a two-stage generative architecture that separates global shape synthesis from part-level decoding. We demonstrate that the resulting assets can be directly integrated into game engines and driven by animation and behavior scripts without manual post-processing.
\end{abstract}



\begin{teaserfigure}
\centering
\includegraphics[trim={0 1mm 0 1mm}, clip, width=\linewidth]{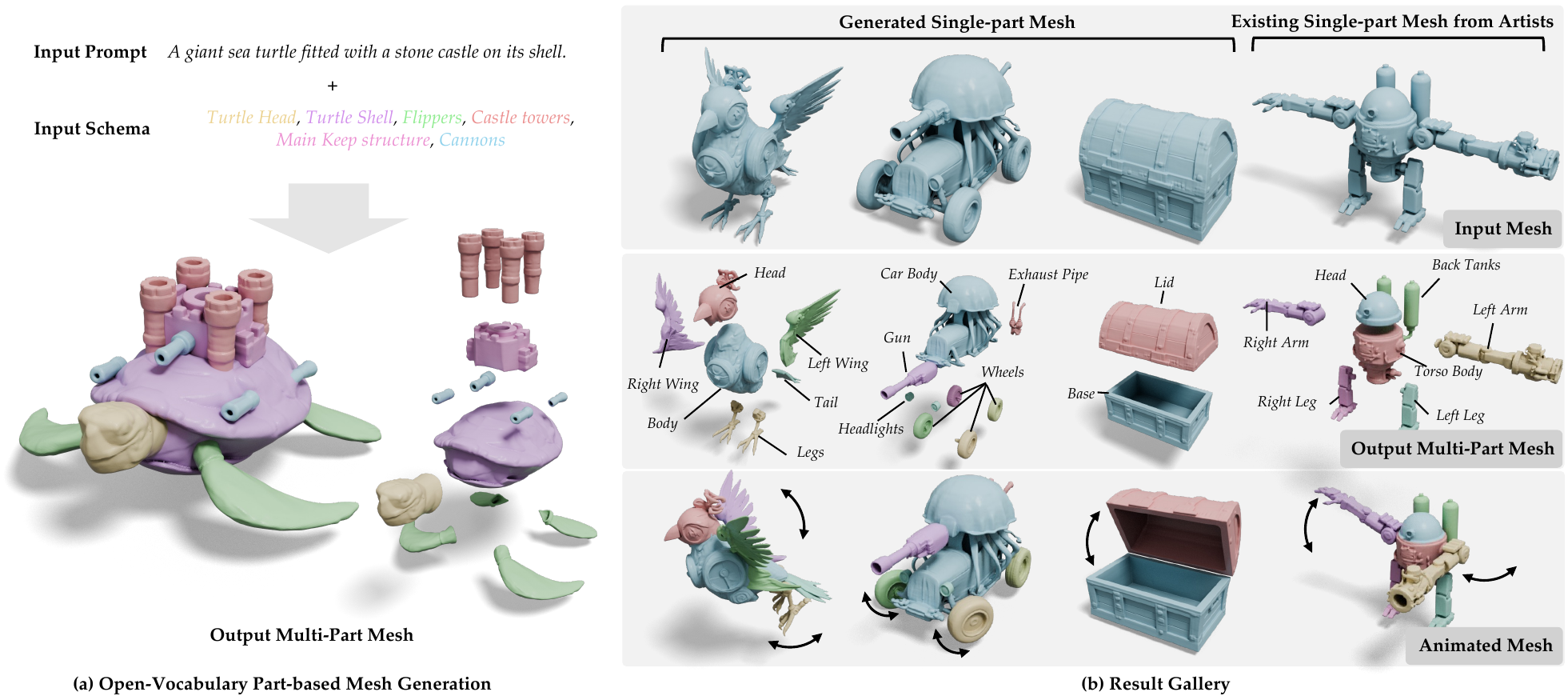}
\caption{\textbf{We propose CubePart, an open-vocabulary part-controllable 3D generator.} (a) Given a text prompt and a schema defining part decomposition, CubePart synthesizes a multi-part 3D object where each component is a distinct, structurally complete mesh. (b) This controllable part-based generative framework directly facilitates the usage of the resulting assets for scripted or physically simulated behaviors (bottom row). Additionally, CubePart can accept an existing mesh as input, decomposing it into semantic multi-part meshes according to the input part schema (last column).
Please refer to the video for animated visualizations.
}
\lblfig{teaser}
\end{teaserfigure}

\maketitle
\renewcommand{\shortauthors}{Zhu et al.}

\section{Introduction}
3D assets in modern games and interactive applications are rarely static. Vehicles require rotating wheels, characters must articulate, containers need to open and close, and many objects respond to physics or scripted events. In game engines, these behaviors are governed by simulation systems, animation rigs, and interaction scripts that operate on a pre-defined set of parts. For an asset to be functional, its mesh must be decomposed into specific semantic components that match the "schema" expected by the game's code. %

Creating meshes that conform to a target part composition remains a largely manual process. Artists must decompose geometry into parts, assign consistent labels, and ensure that the resulting meshes assemble cleanly---an effort that scales poorly with asset diversity. While recent advances in 3D generative modeling have enabled the synthesis of complex geometries from text or image prompts, these methods either produce monolithic meshes without any explicit part structure~\cite{trellis,trellis2,hunyuan3d-2-1} or an arbitrary set of parts~\cite{partpacker,partcrafter}; the user has no control for aligning the resulting parts with the schema required by downstream game logic. For a developer with a game that specifically expects a car to be composed of four wheel parts and one chassis part, a model that generates a random set of part segments is as unhelpful as a model that generates a car as a single monolithic object.

One might attempt to obtain part-level control through 2D grounding, for example, by using an image segmentation model \cite{sam,sam3} to generate segmentation masks for mask-conditioned 3D part generation models like OmniPart~\cite{omnipart}. However, a 2D mask cannot represent or control parts that are hidden from the input view. For instance, the rear tail of an animal cannot be specified or controlled from a single front-facing view. More fundamentally, 2D control signals are view-dependent and ambiguous when lifted to 3D, making them ill-suited for defining complete semantic decompositions of 3D objects.

These limitations highlight a critical need for a 3D-native, schema-driven control interface that allows users to explicitly specify the semantic structure of an object during generation. Such control must also be flexible: different applications may require different decompositions of the same object. For example, one game may need car doors as separate parts to enable opening animations, while another may require the hood to be independently controllable to expose the engine. Fixed or closed-vocabulary part schemas cannot accommodate this diversity of downstream requirements. We argue that text, as a modality, provides a natural and universal interface for such control. Crucially, a text prompt can specify both a global description of the desired object and an explicit \emph{parts schema}, an open-ended list of part names that serves as a structural blueprint for decomposing the object into semantically meaningful components.

In this paper, we present~\methodname, the first generative framework for open-vocabulary, part-controllable 3D mesh generation. Our system takes as input a global text prompt describing the object (e.g., “a jellyfish-themed race car”) together with a desired parts schema (e.g., \{"car body", "front left wheel", …\}). It outputs a set of distinct meshes, one per schema element, that jointly assemble into a coherent object. Because the generation is guided by the user-provided schema, the resulting assets can directly match the requirements of game engines and animation systems without manual intervention (as we demonstrate in Section~\ref{sec:application}).

To support this capability, we introduce \emph{CubePart}, a framework underpinned by a high-fidelity data engine and a novel multi-stage generative architecture. Our data engine leverages vision-language models (VLMs) and a novel 3D-aware "Set-of-Mark"~\cite{setofmark} annotation strategy to curate a semantically grounded dataset of \revised{462K} assets and about \revised{2M} parts. Compared to existing 3D part datasets, ours is both larger scale (\revised{over 11} times larger than PartVerse-XL~\cite{fullpart}) and produces higher quality part labels required for precise open-vocabulary control. Building on this foundation, our architecture employs a two-stage diffusion process: the first stage generates a full mesh conditioned on both the prompt describing the object and the part schema, and the second stage decomposes the full mesh into corresponding parts specified by the schema while ensuring global geometric coherence through a novel cross-part attention mechanism with zero-initialized attention blocks.

In summary, our main contributions include:
\begin{itemize}
    \item A scalable data engine for constructing open-vocabulary, part-labeled 3D datasets from unstructured meshes, leveraging VLMs for 3D-aware clustering and semantic captioning.
    \item A schema-driven two-stage generative architecture that supports open-vocabulary, part-controllable 3D mesh generation while preserving global coherence across parts.
    \item An end-to-end demonstration showing how the generated multi-part meshes can be integrated into game engines and driven by behavior scripts without manual post-processing.
\end{itemize}

\section{Related Work}
\subsection{3D Generative Models}
Recent progress in 3D generative modeling was initially driven by 2D-to-3D lifting approaches, most notably DreamFusion~\cite{dreamfusion}, which introduced Score Distillation Sampling (SDS) to optimize implicit 3D representations using pretrained 2D diffusion priors. A large body of follow-up work~\cite{get3d,magic3d,prolificdreamer,zero123} adopts this paradigm, leveraging strong 2D priors to compensate for limited 3D data. Despite impressive visual quality, these methods rely on view-dependent image supervision and provide only weak constraints on 3D structure, offering no explicit control over semantic part decomposition.

With the availability of large-scale 3D datasets such as Objaverse~\cite{objaverse} and Objaverse-XL~\cite{objaversexl}, 3D-native generative modeling has become increasingly practical. 3DShape2VecSet~\cite{3dshape2vecset} introduces a compact latent-set representation that enables diffusion directly in a 3D-aligned latent space, and subsequent works~\cite{michelangelo,craftsman3d,cube3d,step1x3d,lattice,clay,triposg,hunyuan3d-2-1} scale this paradigm to high-quality, end-to-end 3D asset generation without reliance on 2D distillation. Building on this representation, our method conditions directly on text rather than images, enabling open-vocabulary semantic control over both object appearance and part composition.

A complementary line of work represents 3D shapes using sparse voxel grids to reduce the memory cost of dense voxelization, as in XCube~\cite{xcube}, Trellis~\cite{trellis,trellis2}, SparseFlex~\cite{sparseflex}, Sparc3D~\cite{sparc3d}, and Direct3D-S2~\cite{direct3ds2}. While these methods support localized geometry synthesis and high-resolution detail, they typically generate monolithic meshes and lack explicit mechanisms for semantic part-level control or decomposition.

\subsection{Part-aware 3D Generation} %
The growing demand for structured and interactive 3D assets has motivated research on part-aware 3D generation. Early methods rely on category-specific, part-level supervision, learned through autoencoder-based frameworks such as SPAGHETTI~\cite{spaghetti} and Neural Template~\cite{neuraltemplate}, or diffusion-based approaches including SALAD~\cite{salad} and DiffFacto~\cite{difffacto}. While these methods demonstrate the feasibility of decomposed shape generation, they are restricted to narrow object categories and fixed part taxonomies, limiting their applicability to open-world asset creation and downstream tasks requiring flexible, application-specific part definitions.

Recent methods~\cite{part123,partgen,comboverse} adopt multi-stage pipelines that combine multi-view diffusion–based image synthesis, 2D foundation models for part segmentation, and subsequent 3D reconstruction and composition. Part123~\cite{part123} generates multi-view images from a single input view, applies SAM-based segmentation to extract part masks, and reconstructs parts via multi-view geometry, while PartGen~\cite{partgen} improves robustness by repurposing multi-view diffusion models for multi-view segmentation and part-aware completion. Despite these advances, such approaches remain strongly dependent on 2D segmentation quality and are inherently limited by view-dependent image evidence, often leading to incomplete or inconsistent 3D parts, especially for occluded or self-hidden components.

Several contemporaneous works~\cite{copart,unipart,fullpart,patchalign3d} move toward 3D-native part generation. HoloPart~\cite{holopart} adopts a two-stage pipeline that segments a 3D shape and applies 3D diffusion to complete occluded regions, whereas PartCrafter~\cite{partcrafter} uses a single unified model to directly synthesize multiple 3D parts from an RGB image without pre-segmentation. PartPacker~\cite{partpacker} addresses inter-part contact artifacts via a dual-volume packing strategy in SDF space, while AutoPartGen~\cite{autopartgen} autoregressively generates a variable number of parts with latent diffusion, incurring high computational cost and quality degradation due to error accumulation. 
\revised{BANG~\cite{bang} formulates part generation as an object explosion process and recursive refinement that supports both unconditional generation and various explicit control signals, but often fails to preserve fine-grained geometry due to the lack of explicit per-part supervision.}

To improve part controllability, OmniPart~\cite{omnipart} proposes a two-stage pipeline consisting of a structure planning module that predicts explicit 3D bounding boxes from 2D part masks and images, followed by a 3D-native, spatially conditioned generative model based on Trellis~\cite{trellis} to synthesize 3D parts. X-Part~\cite{xpart} similarly adopts a two-stage design, first leveraging the 3D-native part segmenter P3-SAM~\cite{p3sam} to produce initial segmentations, bounding boxes, and semantic features, and then performing synchronized multi-part diffusion to generate 3D parts.

Despite this progress, existing methods either assume a fixed or learned part vocabulary or infer part structure implicitly from data or 2D segmentation. In contrast, our approach allows users to directly specify an open-vocabulary list of semantic parts at inference time, and guarantees that the generated meshes align with this user-defined structure, enabling direct integration with downstream animation and interaction pipelines.

\begin{figure*}[t!]
\centering
\includegraphics[trim={0 0mm 0 1mm}, clip, width=\linewidth]{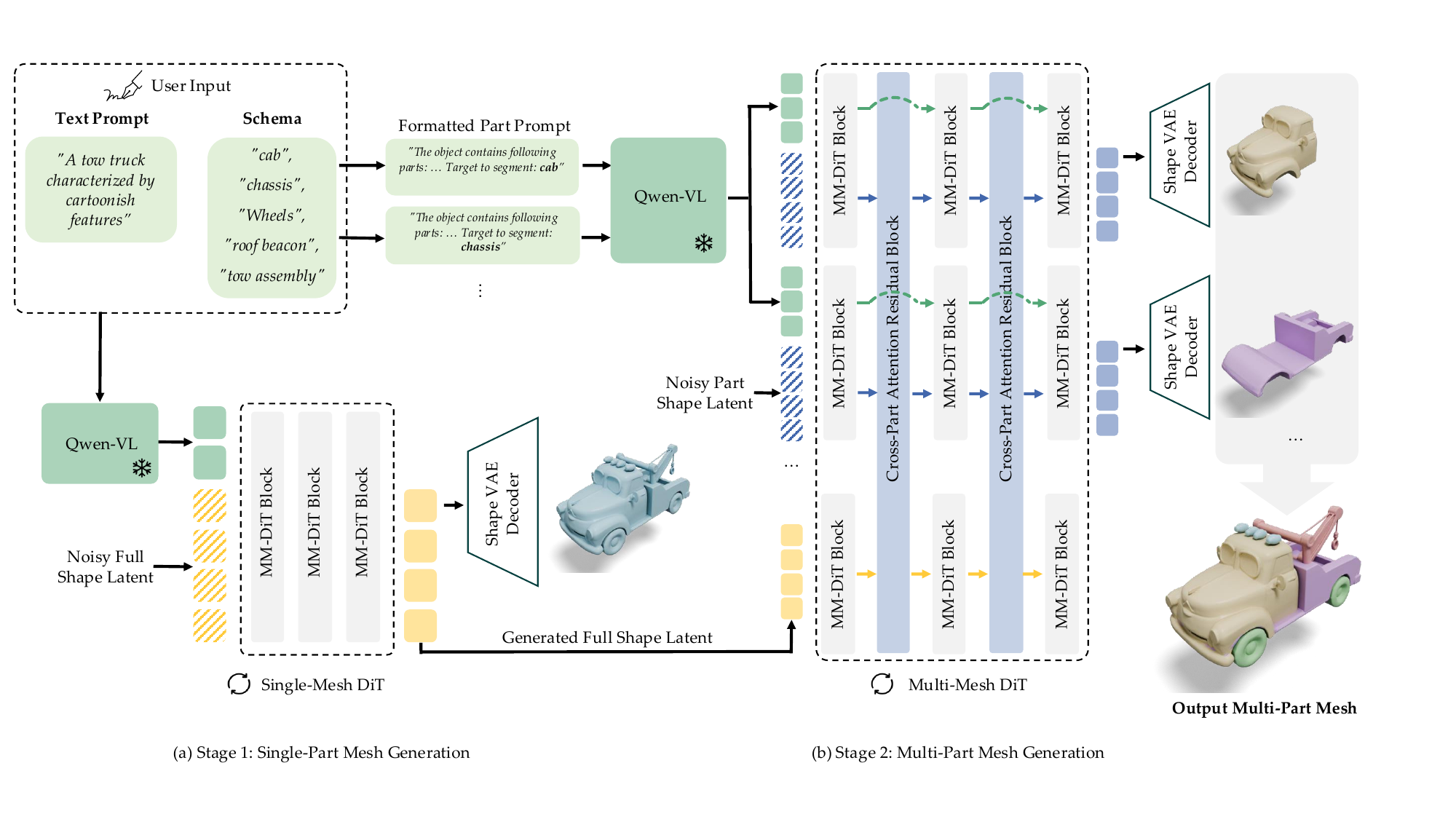}
\caption{\textbf{Overview.} We propose a two-stage framework to generate part-controllable 3D objects conditioned on a global text prompt and a part schema. (a) Single Mesh Generation synthesizes a holistic shape latent using a Multi-Modal DiT (MM-DiT)~\cite{stable-diffusion-3}, conditioned on the prompt and schema encoded by Qwen-VL~\cite{qwen-vl}. (b) Multi-Mesh Generation takes the full shape latent from Stage 1 and decomposes it into distinct part latents. To achieve this, we initialize with the MM-DiT weights from Stage 1 and inject Cross-Part Attention Residual Blocks to enable structural interaction among parts.
}
\lblfig{model_overview}
\end{figure*}

\subsection{3D Part Datasets}
Part-aware generative models rely on datasets in which meshes are decomposed into meaningful components. We define a part dataset as a collection of 3D meshes that are pre-segmented into distinct parts, in contrast to datasets like ShapeNet \cite{shapenet}, ABO \cite{abo} or Objaverse/Objaverse-XL \cite{objaverse,objaversexl} that primarily contain monolithic meshes. These parts generally correspond to meaningful object components (such as the left mechanical arm of a robot), though they may also reflect the structural choices of the original artist. 

We further characterize a part dataset as "open-vocabulary" if each part is paired with free-form natural language descriptions or names, rather than labels drawn from a fixed taxonomy. This contrasts to closed-vocabulary part datasets that enforce a predefined part taxonomy such as ShapeNetPart \cite{shapenetpart} and PartNet \cite{partnet}.

Recent efforts toward open-vocabulary part datasets include PartVerse \cite{copart} and PartVerse-XL \cite{fullpart}, which curate subsets of Objaverse, refine their part segmentation using human experts, and generate part captions using large vision-language models (VLMs). These datasets contain approximately 12k and 40k assets, respectively. PartObjaverse-Tiny \cite{sampart3d} provides manually curated open-vocabulary labels for a uniformly sampled subset of 200 meshes from Objaverse, but is intended primarily for evaluation rather than training. Although these datasets represent important progress, they remain expensive to scale and limited in coverage, motivating automated pipelines that can construct large-scale, open-vocabulary part datasets from unstructured 3D assets.

\section{Open-Vocabulary Part-Controllable 3D Generator}

We aim to generate part-controllable 3D objects conditioned on a global user prompt describing the overall shape, supplemented by a text-based schema that defines its composing parts, \eg a sleek sports car with wheels, door, body, and engine. 
To this end, we propose \methodname, a framework comprised of two key stages: full mesh generation and multi-part mesh generation (\reffig{model_overview}). 
In \refsec{method:preliminary}, we provide a brief overview of the vecset-based diffusion transformer for mesh generation introduced in Craftsman~\cite{craftsman3d} and other follow-up works~\cite{hunyuan3d-2-1,triposg,clay}. 
We then describe how we adapt this architecture to establish our single-part mesh generation pipeline, which generates a full mesh from a user-defined text prompt.
Finally, we introduce the second stage, multi-part mesh generation, detailing how we decompose the single-part mesh into corresponding components defined by the text-based schema.

\subsection{Preliminary: Vecset Diffusion for Mesh Generation}
\lblsec{method:preliminary}

Vecset diffusion models~\cite{craftsman3d,triposg,clay,hunyuan3d-2-1} represent a class of latent diffusion models designed to generate sets of unordered vectors (vecsets) that implicitly encode 3D shapes.
The typical pipeline begins by encoding 3D meshes into latent vector sets using a transformer-based Variational Autoencoder (VAE) using 3DShape2VecSet~\cite{3dshape2vecset}. The VAE decoder employs a Signed Distance Function (SDF) representation, which enables sharper geometry reconstruction.
A diffusion model, often based on flow matching formulations~\cite{stable-diffusion-3}, is then trained on these VAE latents to generate novel 3D shapes from noise. For image-to-3D generation tasks, these models are commonly conditioned on single-view images through visual features, \eg DINOv2~\cite{dino-v2}, injected via cross-attention mechanisms in transformer blocks.

\subsection{Stage 1: Single-part mesh generation}
\lblsec{method:stage1}

While most VecSet diffusion models are image-conditioned, images are not well suited for defining complete 3D semantic structures due to part occlusion. 
To this end, we adapt the VecSet diffusion model for text-to-3D generation.

\paragraph{\textbf{Pretraining.}} To bootstrap the model for more complex tasks, we first pre-train the model on a text-conditioned generation task. We utilize the vecset-based shape VAE~\cite{3dshape2vecset} and adopt the Multi-Modal Diffusion Transformer (MM-DiT) architecture~\cite{stable-diffusion-3}, for text conditioned 3D shape latent generation. 
Additionally, we employ Qwen-VL~\cite{qwen-vl} to encode the text prompt, following Qwen-Image~\cite{wu2025qwenimagetechnicalreport}.
The pretraining dataset consists of approximately 4.7M mesh-text  pairs, combining 745K proprietary assets with about 4M synthetically generated assets for improved text diversity, following the recipe from~\cite{cube3d}. 
This network subsequently serves as the single-mesh generation model to \revised{showcase} an end-to-end pipeline, though our Stage 2 multi-part mesh generation model is able to take any watertight mesh as input.

\paragraph{\textbf{Schema-aware Finetuning.}} While the pre-trained single mesh diffusion model can produce high-quality 3D shapes, the resulting mesh is not guaranteed to contain all the intended parts, even when the input schema is explicitly included in the text prompt. Conversely, certain parts might be disproportionately emphasized (\reffig{ablation_single_mesh}). To address these limitations, we fine-tune the base model on our curated dataset, where the text prompts are structured to explicitly enumerate the constituent parts. The full prompt is: ``\textit{\{global caption\}}. \textit{This object contains the following parts:} \textit{\{list of part labels\}}.''

\paragraph{\textbf{Implementation Details.}} 
To optimize training and inference efficiency, we downsize the original Qwen-Image model. The number of layers is reduced to 21, and the hidden dimension is 1536, which results in 1.9B trainable number of parameters. We adopt the flow matching training objective, following \cite{liu2023flow, ma2024sit, stable-diffusion-3}. During training, given a VAE-encoded shape latent $Z_0\sim\mathcal{D}$ sampled from the training dataset $\mathcal{D}$ and a random noise sampled from the standard multivariate normal distribution $Z_1 \sim \mathcal{N}(0, \mathbf{I})$, the model input latent at timestep $t$ is defined as: $Z_t = tZ_0 + (1-t)Z_1 $ where the timestep $t$ is sampled from a logit-normal distribution and shifted with a factor of 4.0, following \cite{timeshift}. The text condition latent $c$ is obtained from Qwen-VL. The training loss function is defined as:

\begin{equation}
    \mathcal{L} = \qquad  \mathbb{E}_{(Z_0, c) \sim\mathcal{D}, Z_1, t}\ \Vert f_{\theta}(Z_t,t,c) - v_t \Vert^2,
\end{equation}
where $f_{\theta}$ denotes the diffusion network with learnable parameters $\theta$, and $v_t = Z_1 - Z_0$.
We use a batch size of 768 and a learning rate of $10^{-4}$ with a linear warm-up schedule for the first 2,000 iterations. We adopt AdamW as the optimizer, with $\beta$ values set to 0.9 and 0.99, and weight decay disabled.

\subsection{Stage 2: Multi-part mesh generation}
\label{method:stage2}

While Stage 1 has established a robust text-conditioned mesh generator that produces geometry that structurally aligns with the text schema, it produces a single monolithic mesh. In Stage 2, we aim to transform this single mesh into a set of distinct parts. To achieve this efficiently and consistently, we leverage pre-trained weights from Stage 1, adapting the model to output multiple part latents while maintaining the geometric priors learned in pre-training.

\begin{figure}[t]
\centering
\includegraphics[trim={0 2mm 0 0mm}, clip, width=\linewidth]{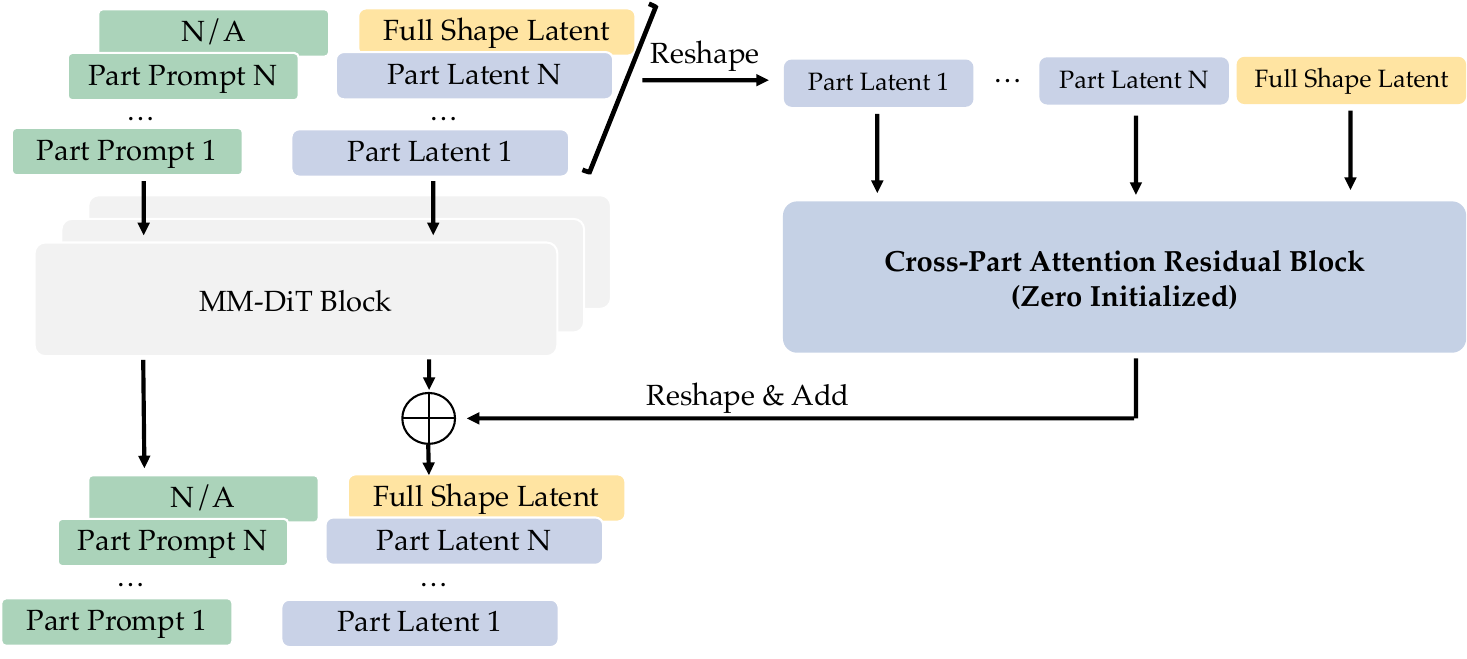}
\caption{\textbf{Cross-part Attention Block.}
A dedicated zero-initialized Transformer block is designed for cross-part global attention. 
The residual block takes in all part latent vectors and the conditional full-shape latent vectors as inputs. We insert this block to facilitate efficient inter-part communication 
while maintaining the pre-trained single-mesh generation capabilities.
}
\lblfig{method_part_attn}
\end{figure}

We represent a multi-part object as a set of $N$ parts $O = \{p_i\}_{i=1}^N$, where each part $p_i$ can be encoded by a set of latent tokens $\mathbf{z}_i = \{z_{ij}\}_{j=1}^K \in \mathbb{R}^{K\times C}$. Here, $K$ denotes the number of tokens per component, and $C$ is the token dimension.

\paragraph{\textbf{Part-aware Prompting.}} A straightforward approach to multi-part generation is to learn a diffusion network $f$ that predicts the latent of a specific part, conditioned on the global context. In this naive baseline, the model takes the form $\mathbf{z}_i = f(\mathbf{z}_{\text{global}}, \text{prompt}_i)$, 
where $\mathbf{z}_{\text{global}}$ is the latent representation of the full mesh. 
To distinguish between different components, we employ a part-aware prompt. 
The text condition is structured as: ``\textit{This object has the following parts:} \textit{\{list of all parts\}.} 
\textit{Target to segment:} \textit{\{target part name\}.}'' 
By explicitly providing the full list of part names, we provide the model with context regarding the other components, helping it better understand the target label and determine its segmentation boundaries.
This prompt guides the model to focus on generating the geometry for a specific semantic part, \eg a ``wheel'' or ``chair leg'', within the context of the whole object.

\paragraph{\textbf{Cross-part Attention Block.}} 
While the straightforward baseline can generate individual components, relying solely on text prompts for global context often results in overlapping or incomplete geometry \revised{(Shown in Table~\ref{tab:multi_part_results})}.
To provide a stronger global context, we must modify the single-mesh model to enable information exchange between parts.
Prior methods like PartCrafter~\cite{partcrafter} and PartPacker~\cite{partpacker} address this 
by altering the original layers of the pre-trained model to perform global attention across all parts. 
However, we empirically found that such extensive modification is unnecessary and can degrade the pre-trained priors. 
Instead, we introduce a dedicated zero-initialized Transformer block specifically for global attention (\reffig{method_part_attn}). 
By inserting this block rather than modifying existing ones, we facilitate efficient inter-part communication 
while minimizing the disruption to the pre-trained single-mesh generation capabilities.

\paragraph{\textbf{Implementation Details.}}
As previously mentioned, we initialize the Stage 2 model using the pre-trained Stage 1 model weights. In total, four cross-part attention blocks are inserted at the 1st, 5th, 9th, and 17th layers. Although these additional blocks increase the model size, we benefit from using fewer computationally expensive cross-part attention blocks while effectively leveraging the pre-trained weights. Training settings generally mirror Stage 1, with the batch size lowered to 72 to handle the multiple parts per sample. \revised{All our diffusion models are trained on 24 H200 GPUs. Stage 1’s training takes about 3 days (1500 GPU-hours). Stage 2’s training time is about 18 hours (450 GPU-hours). Inference on H200 takes ~2-3 seconds for Stage 1, and 3-4 seconds for Stage 2, both including VAE decoding.}

\begin{figure*}[t]
    \centering
    \includegraphics[trim={0 1mm 0 1mm}, clip, width=.8\linewidth]{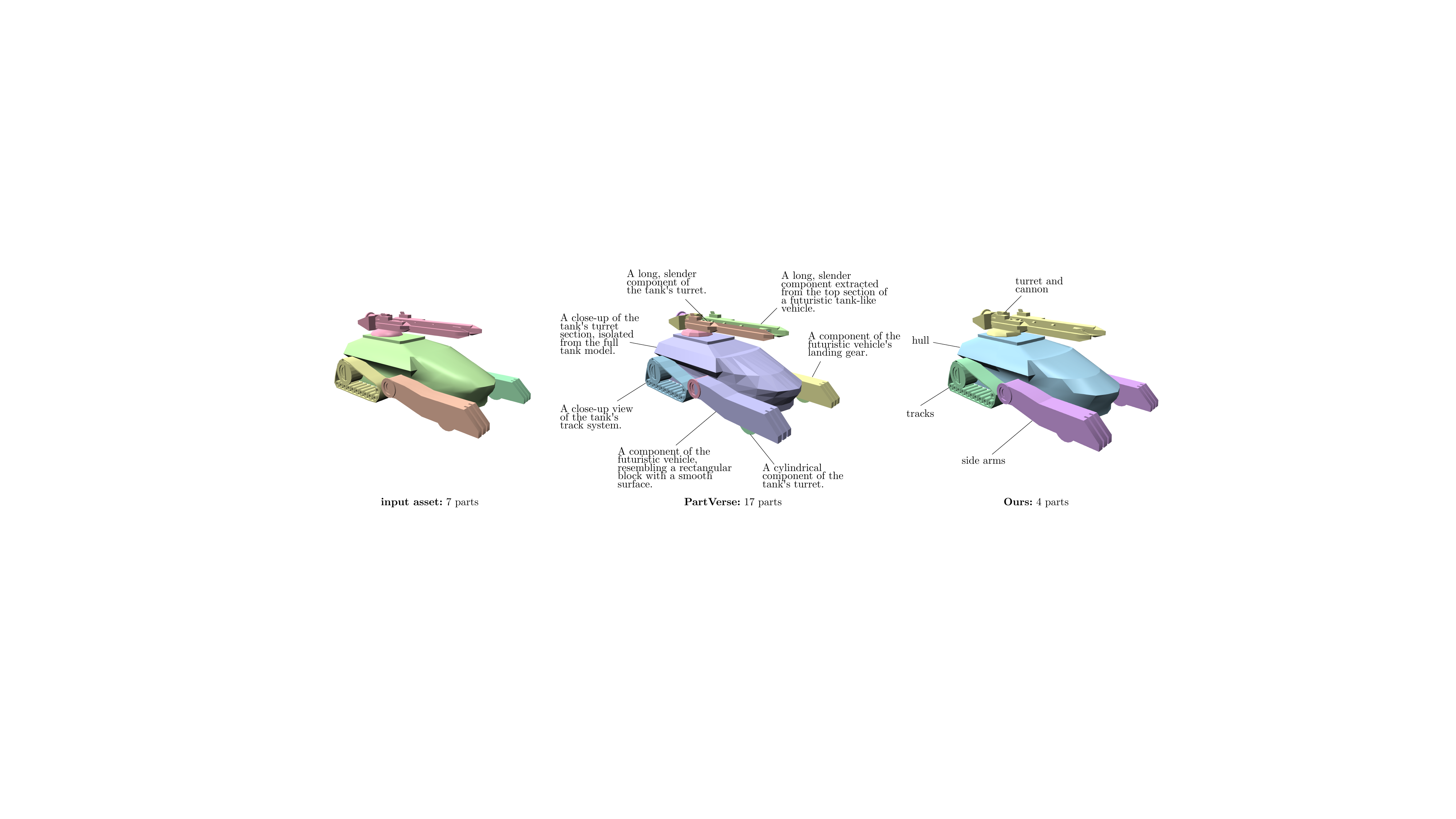}
    \captionof{figure}{\textbf{Part Segmentation and Naming Comparison.} Same Objaverse asset~\cite{objaverse}. \textbf{Top left:} Original artist decomposition (7 parts). \textbf{Middle:} PartVerse~\cite{copart} (17 parts) with VLM captions that exhibit artifacts (``A close-up of ...'') and lack spatial specificity (e.g., ``A red and black circular component of ...''). \textbf{Right:} Our automatic pipeline (4 parts) produces concise, meaningful names (e.g., hull, tracks).}
    \label{fig:partverse_captions_vs_ours_dataset}
\end{figure*}

\begin{table}[t]
    \centering
    \setlength{\abovecaptionskip}{2pt}
    \setlength{\belowcaptionskip}{-2pt}
    \caption{\textbf{Multi-part Training Data.} Composition by source.}
    \label{tab:dataset_sources}
    \small
    \begin{tabular}{@{}l@{\hspace{-6pt}}l@{\hspace{2pt}}l@{\hspace{2pt}}r@{\hspace{2pt}}r@{}}
        \toprule
        Source & Subsets & Content & Assets & Parts \\
        \midrule
        Sketchfab & \makecell[tl]{Objaverse, Texverse,\\PartVerse, PartVerse-XL} & \makecell[tl]{Characters, animals,\\architecture, etc.} & 270K & 1.14M \\
        Commercial & Licensed libraries & Furniture, CAD, etc. & 64K & 201K \\
        Internal & Game collections & Avatars, vehicles, etc. & 129K & 679K \\
        \midrule
        \textbf{Training Total} & & & \textbf{462K} & \textbf{2.02M} \\
        \bottomrule
    \end{tabular}
\end{table}

\begin{table}[t]
    \centering
    \raggedleft
    \setlength{\abovecaptionskip}{2pt}
    \setlength{\belowcaptionskip}{-2pt}
\caption{\textbf{Dataset Comparison.} Scale and annotations vs.\ prior work.}
    \label{tab:dataset_comparison}
    \small
    \begin{tabular}
{@{}l@{\hspace{2pt}}c@{\hspace{2pt}}c@{\hspace{2pt}}c@{\hspace{2pt}}l@{}}
        \toprule
        \textbf{Prior work} & \textbf{Assets} & \textbf{Parts} & \textbf{Open-Voc} & \textbf{Part Text} \\
        \midrule
        ShapeNetPart~\cite{shapenetpart} & 16K & 93k & \xmark & Taxonomy labels \\
        PartNet~\cite{partnet} & 26K & 573K & \xmark & Taxonomy labels \\
        PartVerse~\cite{copart} & 12K & 91K & \cmark & Captions \\
        PartVerse-XL~\cite{fullpart} & 40K & 320K & \cmark & Captions \\
        \midrule
        \textbf{Ours} & \textbf{462K} & \textbf{2.02M} & \cmark & Names \\
        \bottomrule
    \end{tabular}
\end{table}

\section{Dataset}

Training open-vocabulary part-based 3D generation models requires large-scale datasets with descriptive text labels. Current publicly released part datasets fall short. We gathered a new dataset of approximately \revised{462K assets and 2.02M parts}, built using an automated data engine that combines artist-provided segmentations with VLM priors. Using multi-view renders with Set-of-Mark~\cite{setofmark} overlays, we cluster parts and assign semantic names.

Table~\ref{tab:dataset_sources} summarizes our training dataset by source, aggregated from Sketchfab-sourced datasets~\cite{objaverse,texverse,copart,fullpart}, and commercial and internal sources. We deduplicate overlapping sources by asset name, \revised{restrict to permissively-licensed assets,} and exclude assets in PartObjaverse-Tiny~\cite{sampart3d} to prevent test set contamination.

\subsection{Multi-Part Dataset Pipeline}
\label{sec:dataset_pipeline}

Our data engine processes assets through four stages: (1)~preprocessing to filter degenerate geometry and retain assets with 2--32 parts; (2)~VLM-based quality filtering to remove defective meshes and scan artifacts; (3)~VLM-based part clustering and naming; and (4)~postprocessing for watertight mesh conversion and point sampling. Additional details are provided in the supplemental material.

The core of our pipeline is the VLM-based clustering stage. Many 3D assets contain over-segmented parts with inconsistent names. We leverage a VLM to simultaneously cluster related parts and assign semantically meaningful names---for example, merging a car's separate \emph{rim}, \emph{tire}, and \emph{hub} meshes into \texttt{front left wheel} and \texttt{front right wheel}.
This stage operates only on existing part boundaries, without attempting to subdivide parts that are already fused in the source mesh.

\begin{figure}
    \centering
    \includegraphics[width=0.4\linewidth, trim={480pt 370pt 530pt 200pt}, clip]{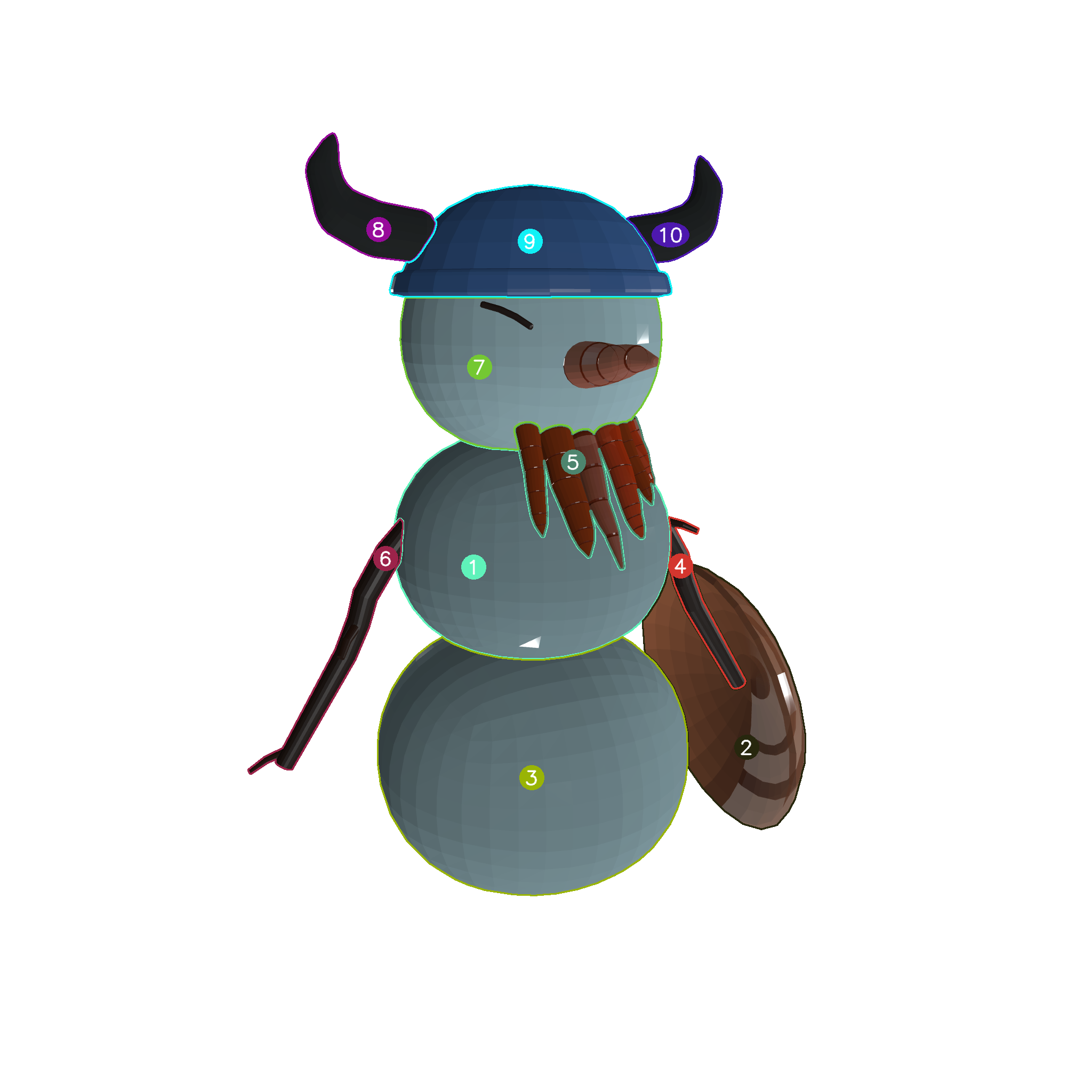}
    \includegraphics[width=0.4\linewidth, trim={480pt 370pt 530pt 200pt}, clip]{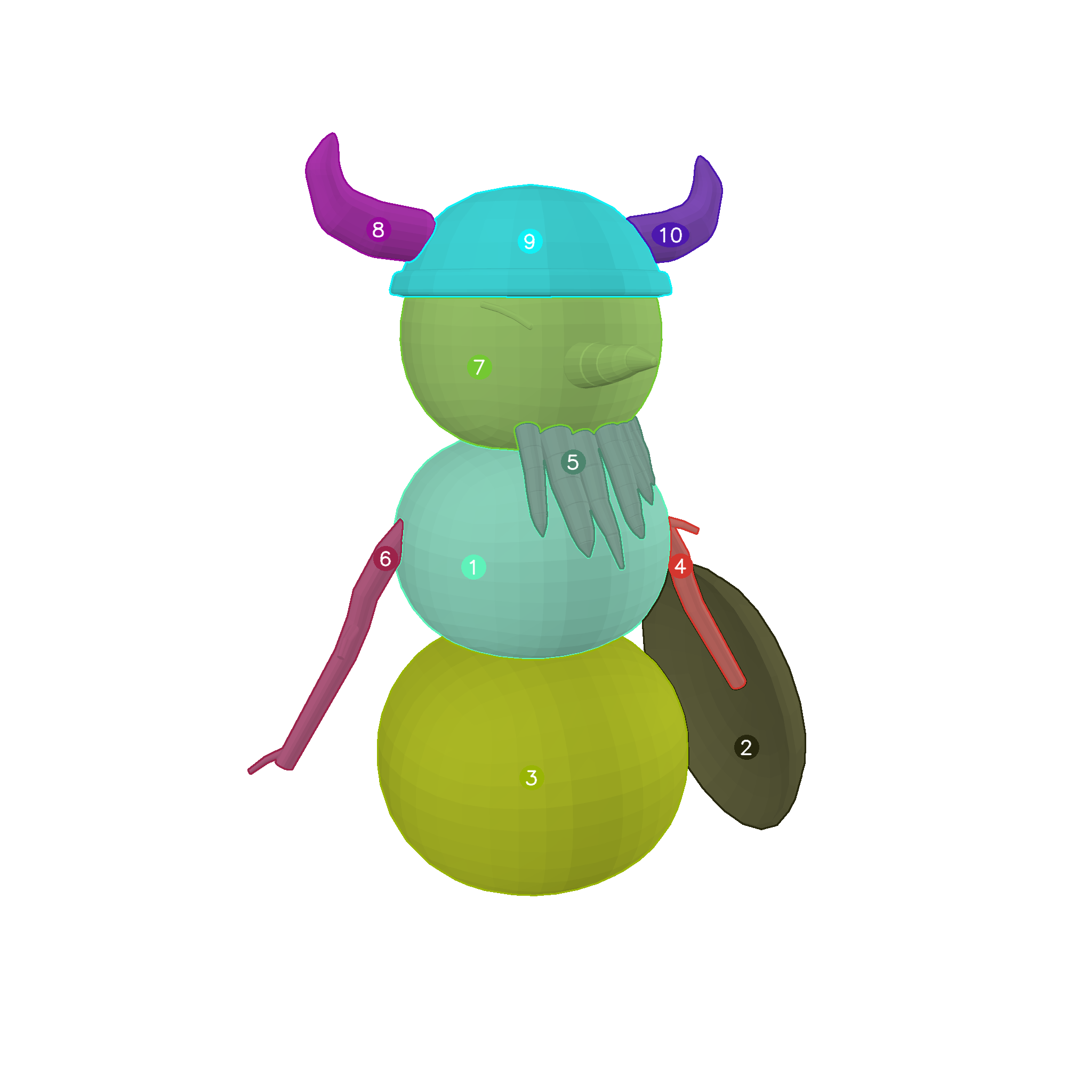}
    \captionof{figure}{\textbf{Set-of-Mark Paired Rendering.} Example pair: textured render (left) with part contours and numbered markers, and part-colored render (right). These paired views are input to the VLM for clustering and naming.}
    \label{fig:vlm_clustering_examples}
\end{figure}

\paragraph{\textbf{Set-of-Mark Rendering.}} Inspired by the Set-of-Mark (SoM) prompting technique~\cite{setofmark}, which enables visual grounding in VLMs by overlaying numeric identifiers on image regions, we adapt this approach to 3D part annotation. Our adaptation differs from the original 2D method in these key ways: we apply SoM to multi-view renders of 3D assets where parts are defined by mesh structure rather than image segmentation, and we generate paired images per viewpoint to provide complementary information.

For each asset, we render 14 orbital viewpoints. Each viewpoint produces a \emph{textured render} with part contours and numbered markers, and a \emph{part-colored render} where each part appears in a distinct solid color. The textured view provides semantic context for accurate naming---helping distinguish a \emph{brick chimney} from a \emph{stone column}---while the part-colored view enables unambiguous part identification. The color of each marker matches the part's contour and solid color across both views, helping the VLM associate parts (see Figure~\ref{fig:vlm_clustering_examples}).

\paragraph{\textbf{VLM-Based Annotation.}} We query a VLM (GPT-5) with all view pairs simultaneously, leveraging the model's ability to reason across multiple images. The prompt instructs the model to group parts into semantic clusters based on function or logical relationships, and assign each cluster a concise, descriptive name. Parts may remain in singleton clusters when they already represent coherent semantic units. The VLM returns structured JSON output containing cluster names  and constituent parts; we resolve edge cases such as duplicate assignments programmatically.

\paragraph{\textbf{Postprocessing.}} After clustering, we convert each part to a watertight mesh using Dual Marching Cubes on a $512^3$ unsigned distance field, then sample surface points with normals for training. \revised{More details about the dataset pipeline are in the appendix.}

\subsection{Comparison}
\label{sec:comparison}

Table~\ref{tab:dataset_comparison} compares our dataset to existing 3D part datasets. The most closely related efforts are PartVerse~\cite{copart} and PartVerse-XL~\cite{fullpart}. Our dataset differs in three main aspects:

\paragraph{\textbf{Scale.}} With 462K assets and 2.02M parts, our dataset is over 11$\times$ larger than PartVerse-XL (40K assets, 320K parts), enabling broader coverage of open-vocabulary part semantics.

\paragraph{\textbf{Automation.}} PartVerse relies on human annotators to correct segmentations, limiting scalability. Our pipeline is fully automatic, using a VLM with Set-of-Mark prompting to cluster parts into functionally meaningful groups (e.g., \texttt{tracks}, \texttt{front axle}). This enables scaling to hundreds of thousands of assets without manual intervention (see Figure~\ref{fig:partverse_captions_vs_ours_dataset}).

\paragraph{\textbf{Part naming vs.\ captioning.}} Our pipeline produces concise part \emph{names} rather than descriptive \emph{captions}. Concise names are more likely to match the part descriptions users provide when querying the model. Caption-based approaches like PartVerse often exhibit VLM artifacts (e.g., ``A close-up of ...'') and assign identical captions to identical parts, providing no way to distinguish them. Our pipeline generates concise names and appends positional adjectives only when needed (e.g., \texttt{left arm}, \texttt{right arm}).

\section{Evaluations}

\begin{figure*}[t!]
\centering
\includegraphics[trim={0 0mm 0 0mm}, clip, width=.99\linewidth]{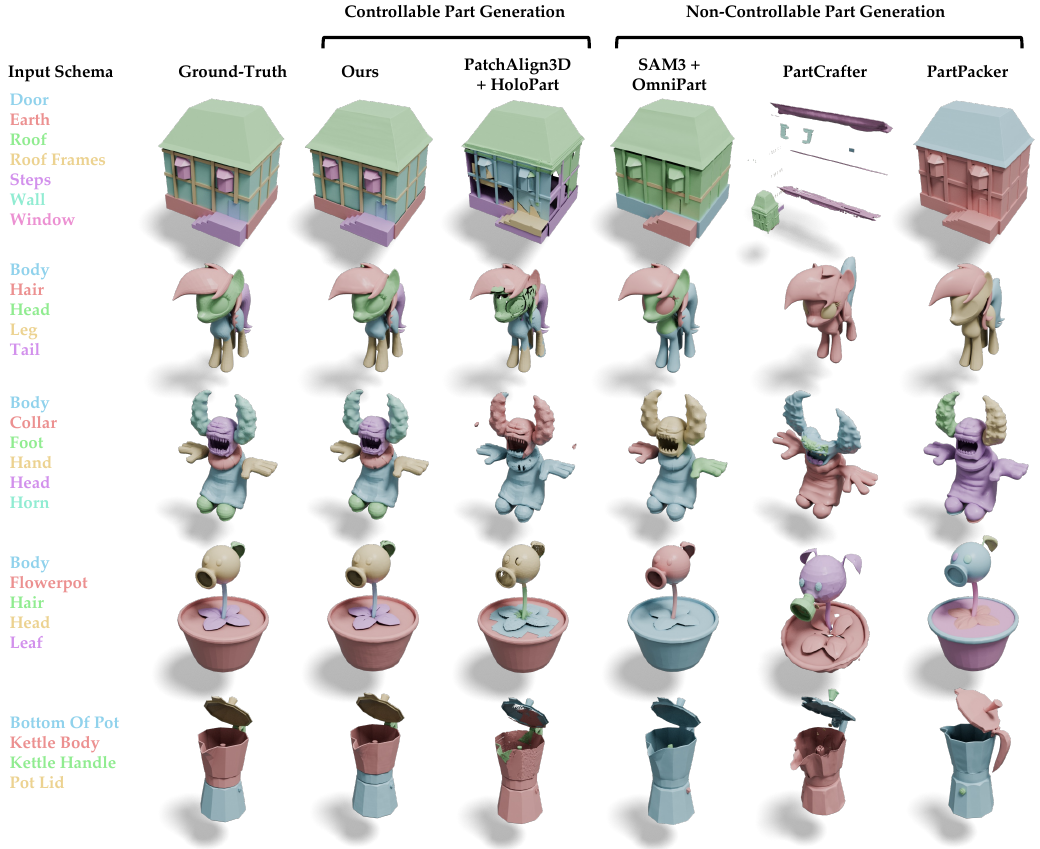}
\caption{\revised{\textbf{Qualitative Comparison of Multi-part Mesh Generation.} We evaluate our method against the two-stage baseline ``PatchAlign3D~\cite{patchalign3d} + HoloPart~\cite{holopart}'' and other image-based part generation methods. Note that both our method and the ``PatchAlign3D + HoloPart'' pipeline condition on a mesh and part schema, whereas the other baselines condition on a single image (with OmniPart~\cite{omnipart} additionally initialized with ground-truth voxels and requiring a text-conditioned 2D segmentation from SAM3~\cite{sam3}). Under the mesh-conditioned setting, our method outperforms HoloPart in both schema adherence and geometric fidelity. The image-conditioned baselines (OmniPart~\cite{omnipart}, PartCrafter~\cite{partcrafter}, PartPacker~\cite{partpacker}) fail to offer user-defined part control and produce noisier segmentation boundaries than our approach.}}

\lblfig{result_multiMesh}

\centering
\includegraphics[trim={0 1mm 0 1mm}, clip, width=\linewidth]{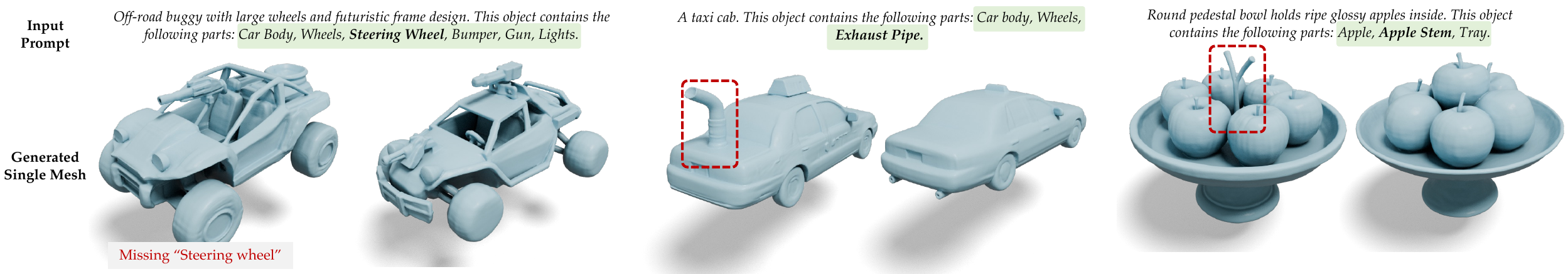}
\captionof{figure}{\textbf{Ablation Study of \revised{Schema-aware} Fine-Tuning.} \revised{In each pair, the left mesh shows generation results without schema-aware fine-tuning, and the right mesh shows our results.} Without fine-tuning, the model fails to include all schema parts (e.g., missing ``Steering Wheel'') or incorrectly emphasizes others (e.g., ``Exhaust Pipe''). Our fine-tuning ensures all requested parts are present and correctly generated.} %
\lblfig{ablation_single_mesh}
\end{figure*}

\begin{figure*}[t!]

\centering
\includegraphics[trim={0 2mm 0 1mm}, clip, width=\linewidth]{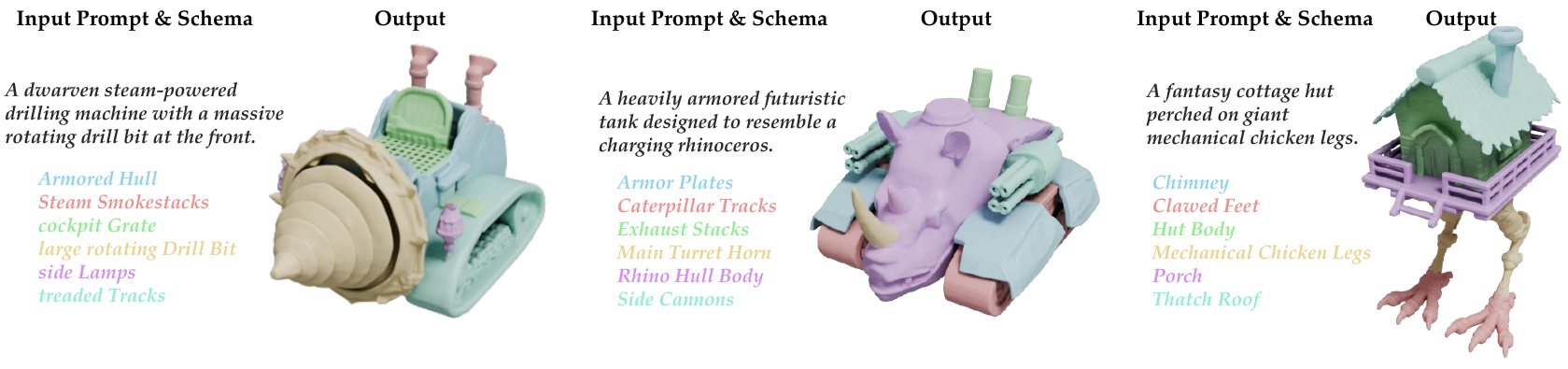}
\captionof{figure}{\textbf{Qualitative Results of Two-Stage Generation.} We present examples generated by our full pipeline. Conditioned on a text prompt and part schema, our method synthesizes detailed global shapes and decomposes them into independent, structurally complete part meshes that adhere to the defined schema.}
\lblfig{result_overall}
\end{figure*}

We evaluate both stages of our pipeline, assessing their ability to generate high-quality meshes aligned with the input part schema.

\subsection{Single-part Mesh Generation}

\begin{table}[t]
    \centering
    \raggedleft
    \setlength{\abovecaptionskip}{2pt}
    \setlength{\belowcaptionskip}{-2pt}
    \setlength{\tabcolsep}{3pt}  %
    \caption{\textbf{Evaluation on Part-based Multi-mesh Generation.} We evaluate the multi-mesh results using CD and F-scores for both individual parts and holistic shapes (formed by concatenating parts). Our method demonstrates consistent improvements in structural completeness and part-level accuracy.}
    \label{tab:multi_part_results}
    \small
    \begin{tabular}{lcccc}
        \toprule
        \multirow{2}{*}{\textbf{PartObjaverse-Tiny}} & \multicolumn{2}{c}{Part-Level} & \multicolumn{2}{c}{Holistic-Level} \\
        \cmidrule(lr{0.5em}){2-3} \cmidrule(lr{0.5em}){4-5}
        & CD $\downarrow$ & F-score $\uparrow$ & CD $\downarrow$ & F-score $\uparrow$ \\
        \midrule
        PartCrafter   & 0.493 & 0.290  & 0.272 & 0.552 \\
        PartPacker   & 0.374 & 0.475  & 0.164 & 0.792 \\
        \midrule
        PatchAlign3D + HoloPart  & 0.309 & 0.549  & 0.050 & 0.970 \\
        \revised{SAM3 + OmniPart} & \revised{0.309} & \revised{0.630} & \revised{0.053} & \revised{0.970} \\
        \midrule
        Ours w/o pre-training & 0.287 & 0.625  & 0.051 & 0.970 \\
        Ours w/o Cross-Part Attention & 0.433 & 0.398 & 0.148 & 0.792 \\
        \revised{Ours w/ PartCrafter-style attention} & \revised{0.386} & \revised{0.529} & \revised{0.089} & \revised{0.864} \\
        
        Ours  & \textbf{0.251} & \textbf{0.743} & \textbf{0.048} & \textbf{0.974}  \\
        \bottomrule
    \end{tabular}
    \lbltbl{eval_multi_mesh}
\end{table}

\begin{figure}[t]
    \centering
    \includegraphics[width=\linewidth, trim={0 13pt 0 15pt}, clip]{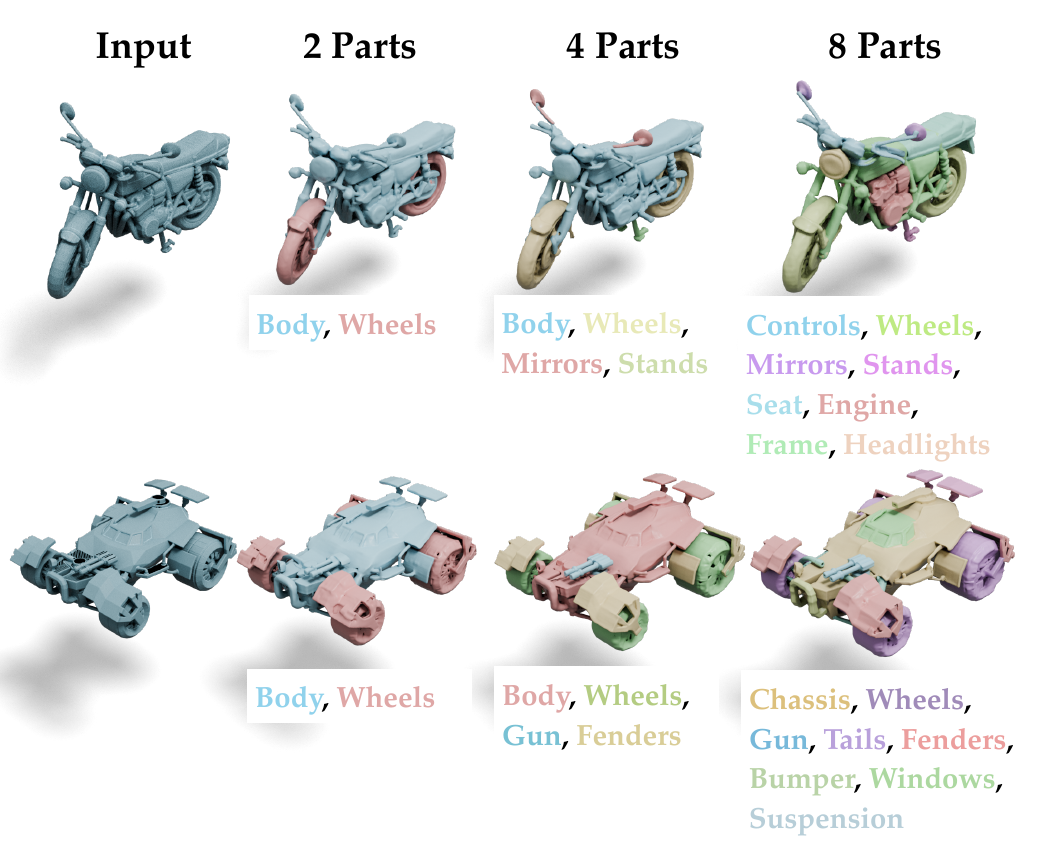}
\caption{\textbf{Qualitative results with varying part schema.} We test our model with  different number of parts on the same input assets. Our method can control generation parts accurately with small components like motorcycle stands and ambiguous closely connected components like chassis and windows. With 2 parts, fenders merge into wheels; with 4 parts, explicit ``fenders'' resolves this ambiguity.}
    \lblfig{result_different_schema}
    \label{fig:result_different_schema}
\end{figure}

 To  verify the effectiveness of our proposed \revised{schema-aware} fine-tuning, we conduct an ablation study. As illustrated in \reffig{ablation_single_mesh}, the pre-trained single-mesh model suffers from missing or incorrectly emphasized parts. With our proposed \revised{schema-aware} fine-tuning, our method resolves these issues, ensuring all requested parts are present and correctly generated.

 Crucially, the Stage 1 single-part mesh generation also serves as pre-training for the Stage 2 multi-part generation, which leads to significant improvement of final output quality both at the part level and holistic level (as shown in Table~\ref{tab:multi_part_results}). By starting from a model that already understands the global geometric prior of a "jellyfish car" or "drone," the second stage can dedicate its capacity to learning inter-part boundaries and part-specific geometries rather than learning basic 3D structure from scratch.

\subsection{Multi-part Mesh Generation}

\paragraph{\textbf{Baselines.}} \revised{We compare Stage 2 of our method with HoloPart~\cite{holopart}, OmniPart~\cite{omnipart}, PartCrafter~\cite{partcrafter}, and PartPacker~\cite{partpacker}. A critical distinction among these methods is their capacity for ``Controllable Part Generation,'' which refers to methods capable of explicitly specifying both the number and the semantic identities of parts. Note that unlike our method, these baselines cannot directly take a text part schema as input. PartCrafter and PartPacker fall short of this definition as they can only control the number of generated parts. HoloPart requires a segmented 3D shape, so we provide it with a 3D segmentation produced by PatchAlign3D~\cite{patchalign3d}, given the ground-truth shape and part schema as input. OmniPart takes a 2D segmentation map as input, for which we provide a text-conditioned segmentation produced by SAM3~\cite{sam3}, and we additionally initialize OmniPart with ground-truth voxels. However, because its resulting 3D parts are not guaranteed to align with this condition in either part quantity or semantic identity, we treat it as a non-controllable baseline, similar to PartCrafter and PartPacker.}

\begin{figure*}[t!]
    \centering
    \caption*{
    \raggedright
    \textbf{Schema}\hspace{.5cm}
    \textbf{Input}\hspace{.5cm}
    \textbf{Output}\hspace{6.7cm}
    \textbf{Behaviors}\hspace{4.5cm}
    }
    \includegraphics[trim={0 12mm 0 0mm}, clip, width=\linewidth]{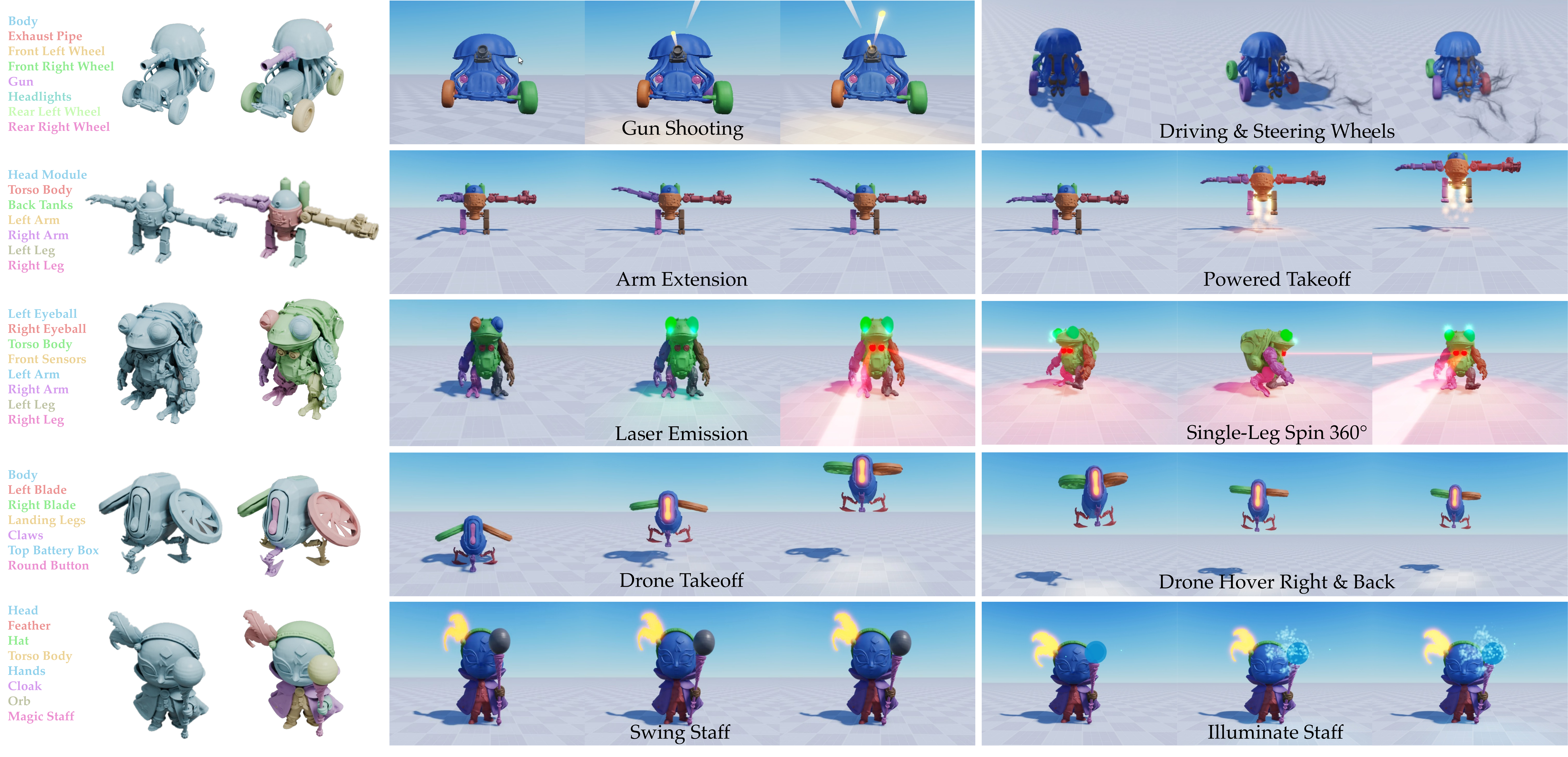}
    \captionof{figure}{\textbf{Application: Applying object behaviors to generated 3D objects.}
    Given an input schema and single-part meshes, our Stage~2 model decomposes the input meshes into parts following the specified schema.
    We then apply object behaviors, including dynamic motions and visual effects. }
    \label{fig:behaviors}
\end{figure*}

\paragraph{\textbf{Comparisons.}} We use PartObjaverse-Tiny as our evaluation dataset.
We employ two geometric metrics to measure shape quality: Chamfer Distance (CD) and F-score. We report these metrics at both part level and holistic (whole mesh) level. All shapes are first normalized to $[-1, 1]$ unit box at holistic level, and F-score is calculated with 0.1 threshold.
For part-level evaluation, we compute CD and F-score for each independent part mesh and average the scores across all parts of an object.
Note that baselines except HoloPart generate an unordered set of parts, requiring us to match the output parts with the inputs for part-level evaluation. 
To do this matching, we use a greedy approach: iterate through the ground truth part prompts, matching each one with the best-scoring output part.

As shown in Table~\ref{tab:multi_part_results}, our method outperforms all baselines consistently across both holistic and part-level metrics. Qualitative results in \reffig{result_multiMesh} further illustrate our method's advantage in decomposing monolithic meshes into parts according to the given schema. Furthermore, Figure~\ref{fig:result_different_schema} highlights that by varying the input schema for a single mesh, our method can accurately modulate both the semantic identity and granularity of the generated parts.

\revised{
Finally, we validate our architectural design choices through an ablation study. Removing the cross-part attention mechanism entirely ("Ours w/o Cross-Part Attention") leads to a severe drop in part-level accuracy, highlighting the critical role of inter-part communication for resolving geometric boundaries. Furthermore, attempting to achieve this communication by modifying pre-trained local attention layers ("Ours w/ PartCrafter-style attention") following PartCrafter~\cite{partcrafter}'s style disrupts the model's learned priors, resulting in noticeably worse performance than our dedicated zero-initialized blocks. We also observe that omitting Stage 1 pre-training reduces overall structural completeness, confirming its value in establishing strong global geometric priors before part decomposition. We also show our end-to-end two-stage pipeline results in \reffig{result_overall}.
}

\section{Application: Generating 3D Objects with Behaviors }
\label{sec:application}

A primary motivation for schema conditioning is to enable behavior-driven generation of 3D objects, where part structures are designed to be compatible with scripted behaviors in interactive 3D environments.
In our workflow, all experiments are conducted in a large gaming platform, where object behaviors are implemented as Lua scripts that directly control individual object parts. We show behaviors in Figure~\ref{fig:behaviors} and the demo video, 
\revised{and provide more details about the behavior script pipeline in the appendix.}

 \paragraph{\textbf{Driving.}} We use the jellyfish car in Figure~\ref{fig:behaviors} to illustrate how part segmentation supports different vehicle behaviors. For basic driving functionality, we specified 5 parts: \textit{body, front left wheel, front right wheel, rear left wheel, rear right wheel}. %
 To add headlights and exhaust effects, we refined the schema to separate \textit{jellyfish body, headlights, exhaust pipe} in addition to the four wheels. Finally, adding a shooting behavior required further segmentation of a {gun} part, along with scripts for bullet generation and visual effects.

\paragraph{\textbf{Characters.}} We evaluate our approach on a diverse set of character types with varying structural and behavioral complexity. The robot character in Figure~\ref{fig:behaviors}, where independently labeled parts such as the head, arms, and weapons enable behaviors such as arm extension and powered takeoff.
The humanoid frog character with articulated limbs and sensory parts support behaviors such as laser emission and single-leg spinning motions. Finally, for the wizard character with external props (e.g., a magic staff), separating the character body from the props allows coordinated motions and visual effects, such as staff swinging and illumination.

\paragraph{\textbf{Flying.}} The drone example in Figure~\ref{fig:behaviors} illustrates part-level control to achieve complex flight behaviors. 
The two propellers are treated as separate parts, allowing for asymmetric actuation for takeoff, hovering, and directional motion. 
In addition, the drone body is segmented into functional components such as landing gear, body shell, and lights, allowing visual effects (e.g., blinking lights) to be applied to individual parts.

\begin{figure}[t]
    \centering
\includegraphics[trim={0 9mm 0 7mm}, clip, width=\linewidth]{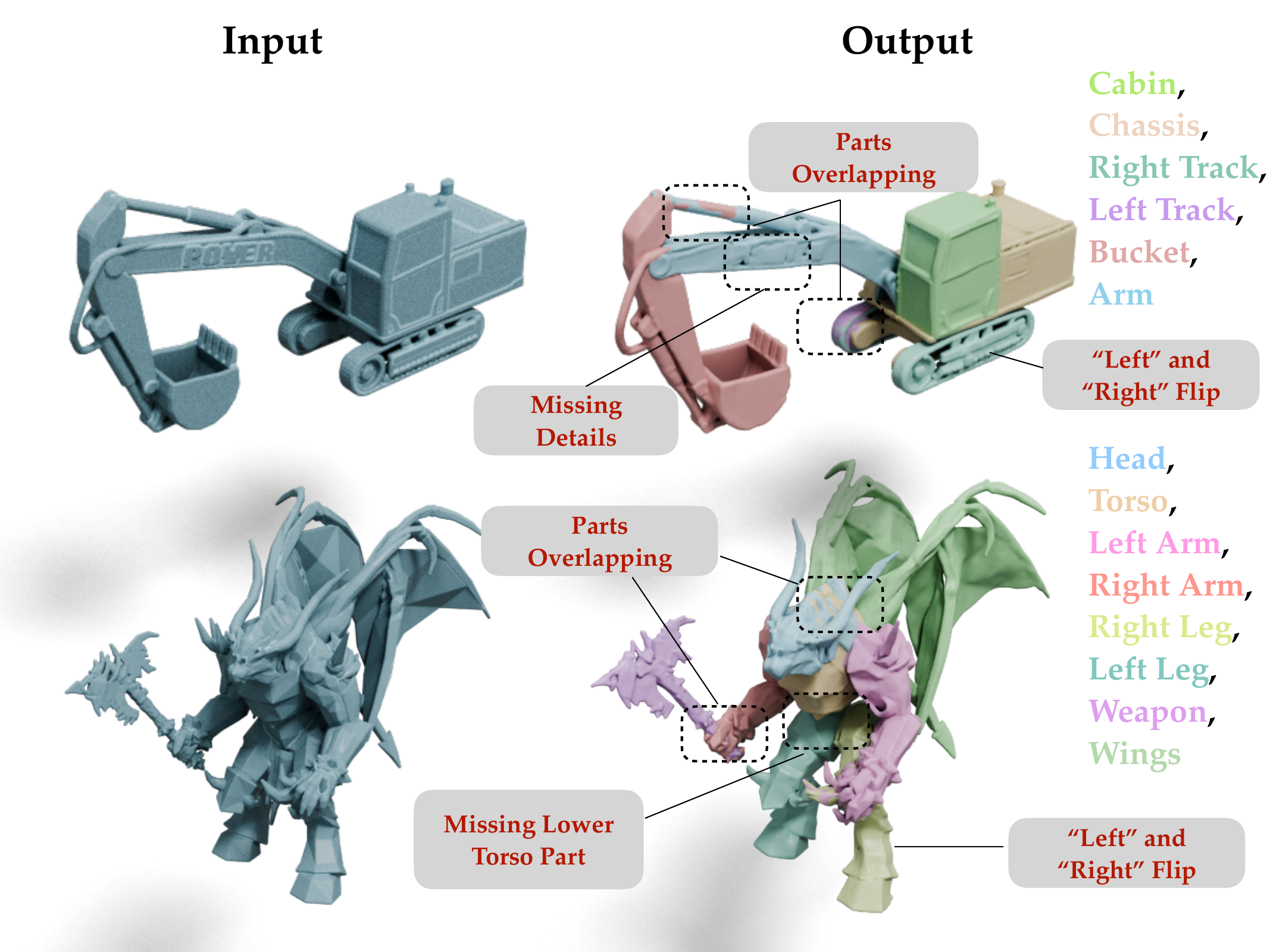}
\caption{\textbf{Failure examples.} Here we show a few typical failure cases where parts can overlap at contact points. The model sometimes can misunderstand spatial relationships as "Left" and "Right", and occasionally drop input components when the input geometry is complicated.}
    \lblfig{result_failure}
    \label{fig:result_failure}
\end{figure}

\section{Limitations and Future Work}
While \methodname represents a significant step towards part-based control of 3D generations to produce game-ready assets, several technical challenges remain.

\paragraph{\textbf{Deformable parts.}} Currently, our model focuses on rigid-body decomposition. While ideal for vehicles, robots, and environmental props, it does not yet support the "skinned" vertex weights required for organic character mesh deformation. Future work could involve predicting skeletal rig weights alongside part geometry.

\paragraph{\textbf{Geometric interpenetration.}} Although our cross-part attention mechanism significantly reduces overlaps, the model can still produce parts that intersect at the boundaries even if the input schema dictates disjoint parts. See Figure~\ref{fig:result_failure} for failure cases.

\paragraph{\textbf{Spatial and Positional Referencing.}} A core difficulty in open-vocabulary part generation is the consistent handling of relative spatial identifiers, such as "front-left" versus "rear-right". While our dataset introduces spatially-aware naming, it inherits the inherent ambiguities of VLM-based labeling. VLMs occasionally struggle with "mirroring" errors, confuse the object's local coordinate system with the camera's view-space, or fail to label occluded parts. This noise in the training data can lead the generative model to occasionally swap symmetrical parts or misplace components along a specific axis (Figure~\ref{fig:result_failure}). Also see the appendix for details.

\begin{acks}
We thank the leadership, Nishchaie Khanna, Karun Channa, Anupam Singh, and David Baszucki, for their support and guidance throughout this work.
We also thank Michael Palleschi, Maurice Chu, Keenan Crane, and Kayvon Fatahalian for helpful discussions.
We are grateful to Zhenyu Zhao, Daniel Chin, Michael Spedden, Alvin Chan, and Saurav Dhakad for setting up the evaluation pipeline as part of the broader project. 
Finally, we are thankful to the ML-Platform team, Anying Li, Yiqing Wang, Steve Han, Sourashis Roy, Chengyi Nie, Wei Zeng, Sal Pathare, Mandar Deshpande, and Andy Shen, for their contributions and collaboration that helped make this project possible.
\end{acks}

\bibliographystyle{ACM-Reference-Format}
\bibliography{main}

\clearpage

\appendix

\section*{Appendix}

In the appendix, we provide more details about our dataset pipeline, additional visual results in \reffig{result_supp}, and animated results in our attached video.

\section{Dataset Pipeline Details}
\label{sec:supp_dataset_pipeline}

This section provides detailed descriptions of each stage in our multi-part dataset pipeline, supplementing the overview in the main paper.

\subsection{Data Sources}
\label{sec:supp_data_sources}

As described in the main paper, our training dataset aggregates assets from multiple sources. The Sketchfab sources (Objaverse, Texverse, PartVerse, PartVerse-XL) overlap significantly; we deduplicate by asset name, prioritizing human-corrected segmentations (PartVerse family) over raw artist segmentations, restrict to permissively-licensed assets, and exclude assets in PartObjaverse-Tiny~\cite{sampart3d} to prevent test set contamination. Commercial and internal sources contribute additional diversity in furniture, architecture, CAD models, and game assets.

\subsection{Preprocessing Stage}
\label{sec:supp_preprocessing}

As the first stage of our data engine, we apply preprocessing steps to prepare assets for annotation. We first remove empty or degenerate geometry parts, then retain only assets containing between 2 and 32 parts. We exclude single-part assets because clustering requires at least two parts, and exclude overly complex assets to maximize the VLM's success rate.

\subsection{VLM-Based Filtering Stage}
\label{sec:supp_vlm_filtering}

Before clustering, we filter assets by quality using a VLM. Each asset is rendered from multiple viewpoints (8 in our configuration) for the model to assess quality. The model identifies mesh defects (tearing, fragmentation), scan artifacts (irregular, noisy surfaces), scene-level content (room sections, object collections, cutaway views), and problematic geometry (zero-volume meshes, overly thin structures). It also outputs complexity scores for geometry and texture, along with a brief description for asset indexing. Based on the identified tags, the VLM classifies each asset into three quality tiers: poor, moderate, or excellent, prioritizing geometric complexity over texture quality and scoring conservatively when borderline. Only moderate and excellent assets proceed to clustering.

\subsection{VLM-Based Clustering and Naming Stage}
\label{sec:supp_vlm_clustering}

This section describes the core annotation stage of our pipeline, which uses a VLM to simultaneously cluster related parts and assign semantically meaningful names.

\subsubsection{Set-of-Mark Rendering}
\label{sec:supp_som_rendering}

Inspired by the Set-of-Mark (SoM) prompting technique~\cite{setofmark}, which enables visual grounding in VLMs by overlaying numeric identifiers on image regions, we adapt this approach to 3D part annotation. While the original SoM method partitions 2D images using segmentation models and overlays marks for tasks such as referring segmentation or phrase grounding, our adaptation differs in three key ways: (1)~we apply SoM to multi-view renders of 3D assets, where parts are defined by the mesh structure rather than image segmentation; (2)~we generate \emph{paired} images per viewpoint---one textured and one colored per part---to provide complementary information; and (3)~since our part masks are derived directly from the 3D geometry and are guaranteed non-overlapping, we place each mark independently at the point with maximum distance to the part boundary, simplifying placement compared to the original approach.

For each asset, we render the scene from 14 orbital viewpoints to ensure comprehensive part visibility. For each viewpoint, we generate a pair of square-proportioned images:
\begin{itemize}
    \item A \textbf{textured render} showing the asset with its original textures, overlaid with part contours and numbered markers (numeric identifiers enclosed in a colored circle).
    \item A \textbf{part-colored render} where each part is rendered in a distinct solid color, with the same numbered markers overlaid.
\end{itemize}

The paired representation provides complementary information: the textured view supplies semantic context essential for accurate naming---for instance, helping distinguish a \emph{brick chimney} from a \emph{stone column}, or a \emph{tiled roof} from a \emph{shingled roof}---while the part-colored view provides unambiguous part segmentation masks that facilitate precise identification and clustering. We found this pairing beneficial through experimentation: using only textured renders resulted in poor clustering accuracy, as parts were harder to isolate visually, while using only part-colored renders led to degraded naming quality due to lost semantic context. The paired approach combines the strengths of both.

Crucially, the color of each numbered marker matches both the part's solid color in the part-colored render and its contour color in the textured render, ensuring visual coherence across both images. This consistent color coding helps the VLM associate parts across the paired views.

\subsubsection{VLM Annotation}
\label{sec:supp_vlm_annotation}

We query a VLM with all view pairs simultaneously, leveraging the model's ability to reason across multiple images using visual tokens. The prompt instructs the model to: (1)~group parts into semantic clusters based on function or logical relationships, rather than visual similarity or spatial proximity alone; and (2)~assign each cluster a concise, descriptive name.

The prompt explicitly allows for \emph{identity clustering}, where a part may remain in its own singleton cluster if it already represents a coherent semantic unit---in which case the VLM effectively provides a name for that individual part.

This stage operates only on the existing part boundaries; it cannot subdivide parts that are already fused in the source mesh. If an asset contains under-segmented parts (e.g., an avatar with a separate head part but a single body part that combines the limbs and torso), the VLM assigns the most descriptive name possible given the combined geometry---it cannot name the arms, legs, and torso individually.

The VLM is instructed to return a structured JSON output containing cluster names and their constituent parts, referenced by the numeric identifiers shown in the Set-of-Mark renders. We handle edge cases through post-processing: duplicate assignments are resolved by keeping each part in its first-assigned cluster, and in rare cases where the VLM groups all parts into a single cluster---effectively collapsing the part structure---the asset is filtered out.

Although source assets may contain existing part names or hierarchical structure, we do not incorporate this metadata into the VLM prompt. Existing names are often noisy and could mislead the model; omitting them also keeps the mechanism general across diverse asset sources without per-dataset adaptation.

We compared GPT-4o and GPT-5 and found that GPT-5 produced more accurate clusters with finer granularity when needed, and more consistent naming conventions, particularly for complex assets with many parts.

\subsection{Postprocessing Stage}
\label{sec:supp_postprocessing}

After clustering and naming, we prepare each asset for training. Artist-created meshes are often non-watertight---containing open surfaces, self-intersections, or non-manifold edges---which prevents reliable inside/outside queries needed for occupancy supervision. We convert each part to a watertight mesh by computing an unsigned distance field on a $512^3$ grid and extracting a level set using Dual Marching Cubes. We then sample 128K visible surface points with normals from each part and 128K visible surface points with normals from the full mesh. Finally, we generate pre-shuffled training and validation epochs, subsampling surface points using Farthest Point Sampling (FPS) to ensure uniform coverage. We apply the same postprocessing for the single-mesh training data.

\subsection{Dataset Limitations}
\label{sec:supp_dataset_limitations}

Our approach has several limitations. Parts that are not visible in any rendered view---either because they are too small or fully occluded---cannot be identified by the VLM and are assigned a special label indicating they remain unlabeled. For under-segmented parts, the assigned name may describe only the most visually prominent component rather than the whole. Positional adjectives such as ``left'' and ``right'' are occasionally confused, and some assets exhibit naming inconsistencies. These imperfections are acceptable trade-offs for a fully automatic pipeline that scales to hundreds of thousands of assets. When higher annotation quality is required, a human-in-the-loop review stage can be applied to filter or correct noisy annotations.

\section{VLM Prompts}
\label{sec:supp_vlm_prompts}

We provide the complete prompts used in our VLM-based pipeline stages, as described in the main paper and in Section~\ref{sec:supp_dataset_pipeline} of this document.

\subsection{VLM Prompt for Quality Filtering}
\label{sec:supp_vlm_filtering_prompt}

The filtering stage uses a VLM to assess asset quality before clustering. The prompt instructs the model to identify visual flaws from a fixed tag vocabulary, assess geometric and texture complexity, and assign an overall quality score. Following the text prompt, we append each of the 8 rendered views with its name. The complete message structure sent to the VLM is the following:

\vspace{0.5em}

\begin{shaded}
{\footnotesize
\begin{verbatim}
You are a visual quality inspector for 3D assets. The images are 
displaying distinct views of a textured mesh. Your task is to 
examine the 3D asset shown in the multi-view rendering and 
determine whether it exhibits any of the following visual flaws 
or characteristics.

Only include tags from the following list. Do not invent or infer 
new tags, even if they seem reasonable. If no tags apply, return 
an empty list. Do not summarize. Return only the matching tags as 
a JSON array.

Here are the tags you may apply (you may select more than one):
### Tag Vocabulary:

- mesh tearing — there is tearing in the mesh, missing polygons, 
  or visible gaps in the geometry that are semantically incorrect 
  (possibly a scanned 3d object)
- 3d scan — mesh geometry appears to be from a 3D scan with 
  irregular, noisy surfaces lacking clean geometric lines, precise 
  edges, or mathematical precision typical of modeled assets
- cutaway view — the mesh appears to be sliced open or missing 
  walls, exposing the interior (e.g. walls of a house removed to 
  show room interiors, section of a ship's hull cut away to 
  display internal compartments, etc). However, natural openings 
  like doorways, windows, or opening that are part of an object's 
  intended design are NOT cutaway view (e.g. the visible interiors 
  of a car from its open windows, the visible cabin of a ships 
  from its doorway, etc.). Those are not cutaway views.
- fragmented object — there is an object composed of two or more 
  disconnected pieces
- multiple objects — the image contains more than one distinct 
  object, not a single unified mesh
- collection of objects — the mesh represents a grouped collection 
  or set of related items (e.g., a toolkit, dinnerware set, or 
  cluster of similar objects)
- mini-scene-like — the mesh is an indoor room or part of an 
  environment, not a standalone object
- room section — the mesh represents architectural elements or 
  room components like wall panels, prefab sections, or modular 
  building parts
- overly complex plants/foliage — the mesh contains intricate 
  plant matter, leaves, branches, or botanical elements with high 
  geometric complexity
- overly thin structures — the mesh contains structures that are 
  extremely thin, wire-like, or have minimal thickness that may 
  cause rendering or processing issues
- no recognizable object — there is no clear or identifiable shape
- heavily occluded views — most of the viewing angles are unable 
  to see semantically meaningful parts of the object (e.g. a 
  stairwell surrounded by walls on three sides)
- zero volume mesh — The object is truly two-dimensional and has 
  **absolutely no measurable thickness** in 3D space. It is a 
  flat surface or an open sheet. Examples: a single polygon card, 
  poster, paper sheet, or an unclosed terrain slice. Exclude all 
  solid objects like swords, knives, rulers, tablet computers, or 
  railroad tracks, which could plausibly exist as a free-standing 
  objects in 3D space. Thin solids (phones, tracks, tablets) are 
  NOT zero volume meshes. Only flat, one-sided polygons qualify.
- has baseplate — the object is situated on top of a base plate 
  (e.g. a thin platform, rocky formation, cutout piece of ground 
  or turf)
- empty image — all views are completely or approximately blank. 
  There is little or no visible content due to rendering failure, 
  bad normalization or missing geometry


### Output format (JSON):
Return a list of all applicable tags.

Return a complexity score for the mesh geometry:
- poor - the asset has very simplistic geometry, with an 
  over-smoothed or blocky characteristic
- moderate - the asset has a normal level of geometric complexity
- high - the asset's geometry is extremely detailed, with high 
  frequency bumps and grooves

Return a complexity score for the texture:
- poor - the asset has a very simplistic texture pattern, uses a 
  very simplistic color pallet, or is generally low resolution
- moderate - the asset has a normal level of textural complexity
- high - the asset's texture is extremely detailed, with high 
  frequency color patterns

Return an overall quality score for the mesh. Prioritize geometric 
complexity and interesting standalone objects over texture quality 
when assigning scores. Take into account the use case when 
assigning a quality score. If an asset's quality is borderline 
between two scores, be conservative and assign the lower score.
- poor — asset is of low quality (3d scan, mesh tearing, not 
  recognizable, fragmented) or asset that is a prefab scene assets 
  (room section, mini-scene-like, multiple distinct objects forming 
  a collection like a car next to another car or two axes - we care 
  about intra part object, not scene level) that should likely be 
  removed from dataset before training. Assets tagged as '3d scan', 
  'multiple objects', 'cutaway view', 'room section', 
  'mini-scene-like', 'overly complex plants/foliage', etc. are of 
  poor quality and should be tagged as 'poor'.
- moderate — asset represents an interesting standalone object, 
  probably good enough for pretraining a large model. Simplistic 
  geometry, simple texture, simplistic design, blocky, low-poly, 
  low-geometric details are still going in the 'moderate', as long 
  as the asset is recognizable and not broken/categorized as 
  'poor'. For instance, a low-poly humanoid character is acceptable 
  and should be tagged as 'moderate'. As rule-of-thumb, no tags 
  are commonly associated with assets scored as 'moderate'.
- excellent — asset has high geometric complexity, represents a 
  high-quality interesting standalone object, and can be used in a 
  small golden training set.

Return a short text description of the textured mesh. This will be 
used for searching the assets, not for training a model. If the 
image is empty or ambiguous, the description can simply be "Empty" 
or "Unknown".

The output should be in json format, specifying the list of "tags", 
the "geometric complexity", the "texture complexity", the 
"reasoning" used to determine the overall quality of the asset, 
the quality "score", and a brief "description" of the asset.

Here is an example output:
{
    "tags": [
        "mesh tearing",
        "heavily occluded views",
        "scene-like",
    ],
    "geometric complexity": "high",
    "texture complexity": "moderate",
    "reasoning": "The asset represents a rocky terrain with dense 
      forest. There are clear gaps in the mesh where there 
      should be terrain, indicating the mesh is incomplete with 
      tears. Many areas of the terrain are not visible from any 
      of the provided views because of the density of the trees. 
      While the asset has very detailed geometry seen in the 
      foliage and rocky portions, and moderately detailed 
      texturing, its scene-like nature and incomplete mesh 
      indicate limited utility for single-object 3D training. 
      Therefore, it deserves a "poor" quality score, and should 
      probably not be included in the training set.
    "score": "poor",
    "description": "An outdoor scene with trees, rocks and dirt.",
}
front_tilt:
[IMAGE: textured render]

front:
[IMAGE: textured render]

right_tilt:
[IMAGE: textured render]

... (repeated for all 8 views)
\end{verbatim}

}
\end{shaded}

\subsection{VLM Prompt for Part Clustering and Naming}
\label{sec:supp_vlm_clustering_prompt}

The clustering stage uses a VLM to group related parts and assign semantic names. The prompt is structured into three components: (1)~a system context that describes the paired image input format, (2)~task instructions including internal reasoning guidelines and clustering rules, and (3)~an output format specification with few-shot examples demonstrating both grouping and identity clustering scenarios.

Following the text prompt, we append each view with its name and the corresponding image pair (textured render followed by part-colored render). The complete message structure sent to the VLM is the following:

\vspace{0.5em}

\begin{shaded}
{\footnotesize
\begin{verbatim}
You are an expert in 3D asset analysis, specializing in part 
identification and semantic grouping. You will receive pairs of 
images for a 3D asset. The first image in each pair shows the 
asset with its original textures and overlays with numbers for 
each part and a contour. The second image shows the same view, 
but with numeric overlays on each part and part contours drawn 
in a single color. This second image is designed to help you 
identify and isolate specific parts.

Your task is to group all visible part IDs into high-level 
semantic clusters based on function, assembly, or logical 
relationship.

## Internal Reasoning Guidelines
To create accurate clusters, you must first mentally identify 
each part. Follow these rules in your reasoning:
  - Identify parts with concise, singular, engineering-style 
    names (e.g., "wheel", "upper arm", "rear bumper").
  - Take some perspective: IDs are centered and can represent 
    a whole body / object.
  - Prefer the most specific commonly used term visible.

## Clustering Rules (Final Output)
  - Create logical, high-level groups. Good clusters represent 
    functional systems (e.g., "propulsion", "suspension") or 
    major assemblies (e.g., "front axle", "front lights").
  - Do not cluster parts solely based on visual similarity or 
    proximity; focus on semantic relationships.
  - Do not cluster parts far away that have no logical 
    connection.
  - Cluster names should be descriptive, often plural or 
    collective.
  - Every visible part ID MUST be included in exactly one of 
    the semantic_clusters.
  - **Identity Clustering:** If the individual parts *already* 
    represent the most logical semantic grouping (e.g., each 
    part is a distinct, high-level component like "engine", 
    "gearbox", "chassis"), then returning each part as its own 
    cluster is the correct and preferred output. Do not force 
    illogical merges.
  - Ensure the clusters are holistic and cover all the 
    identified parts.
  - More complex objects with many parts may require more 
    clusters; simpler objects may need fewer.
  - When using position adjectives (e.g., "front", "rear", 
    "left", "right"), ensure they are accurate based on the 
    object perspective shown in the images and not from the 
    viewer's perspective.
  - Be concise in your cluster naming; avoid unnecessary words. 
    If a single word suffices, use it. Avoid using descriptive 
    phrases unless absolutely necessary for clarity. Use 
    adjectives just to distinguish between similar parts like 
    by location (e.g., "left door" vs. "right door"), but do 
    not add extra descriptive terms. Avoid using the whole 3D 
    asset name in the cluster names.

## Output Format
Respond with a single JSON object ONLY. The object must contain 
a single key: "semantic_clusters".

### Example 1: Grouping components
{
  "semantic_clusters": [
    {
      "cluster_name": "wheels",
      "part_ids": [1, 2, 3, 4]
    },
    {
      "cluster_name": "engine",
      "part_ids": [5, 6]
    }
  ]
}

### Example 2: Identity clustering (already well-grouped)
{
  "semantic_clusters": [
    {
      "cluster_name": "front left wheel",
      "part_ids": [1]
    },
    {
      "cluster_name": "front right wheel",
      "part_ids": [2]
    },
    {
      "cluster_name": "engine block",
      "part_ids": [3]
    },
    {
      "cluster_name": "chassis",
      "part_ids": [4]
    }
  ]
}

front:
[IMAGE: textured render with contours and numbered markers]
[IMAGE: part-colored render with numbered markers]

left:
[IMAGE: textured render with contours and numbered markers]
[IMAGE: part-colored render with numbered markers]

left_tilt_top:
[IMAGE: textured render with contours and numbered markers]
[IMAGE: part-colored render with numbered markers]

... (repeated for all 14 orbital views)
\end{verbatim}

}
\end{shaded}

\begin{figure}[t]
  \centering
  \includegraphics[width=\linewidth, trim={0 0 0 22pt}, clip]{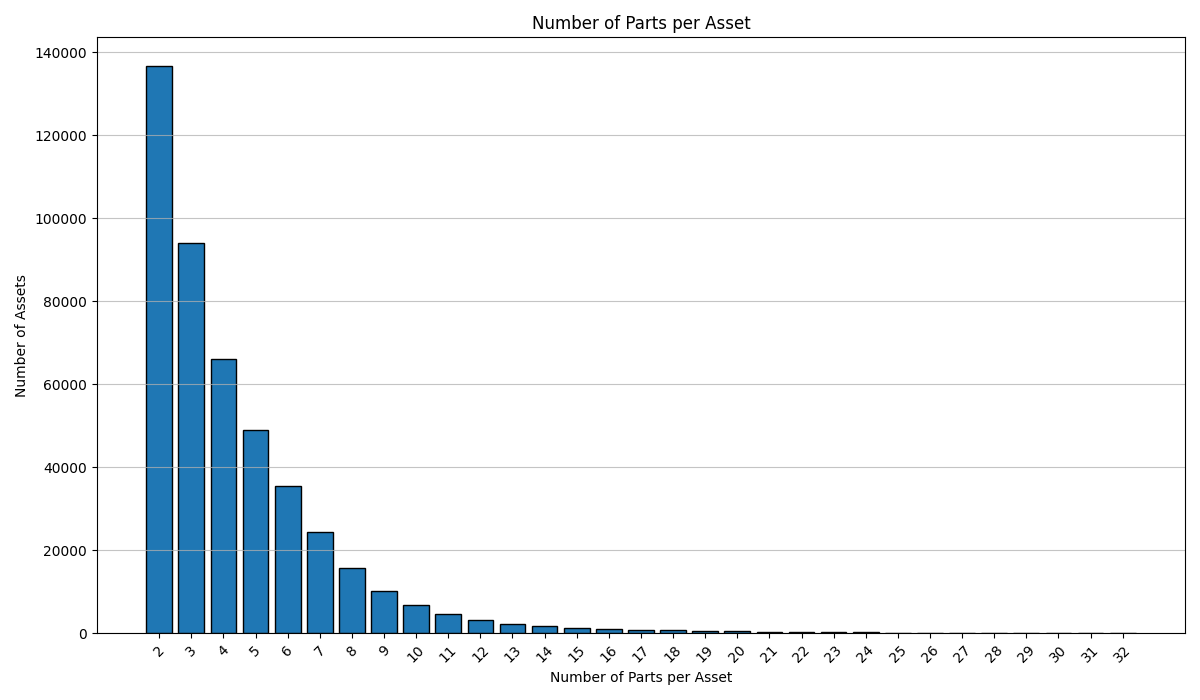}
  \vspace{-20pt}
  \caption{\textbf{Distribution of parts per asset} across the 462K training assets (Sketchfab, commercial, and internal sources combined). The majority of
assets fall in the 2-10 parts range.}
  \vspace{-10pt}
  \label{fig:histogram_datasets}
\end{figure}

\section{Data Distribution}
\label{sec:supp_data_distribution}

Figure~\ref{fig:histogram_datasets} shows the distribution of parts per asset across our 462K training assets. Parts per asset follow a right-skewed distribution: 30\% of assets have exactly 2 parts, 45\% fall in the 3--5 range, 20\% in 6--10, and only 4\% exceed 10 parts (max 32).

\definecolor{codekeyword}{RGB}{0, 0, 180}
\definecolor{codecomment}{RGB}{0, 128, 0}
\definecolor{codestring}{RGB}{163, 21, 21}
\definecolor{codestage}{RGB}{200, 50, 50}
\definecolor{codebg}{RGB}{248, 248, 248}

\lstdefinelanguage{LuauCustom}{
  keywords={local, function, end, for, do, in, if, then, else, return, nil, true, false},
  keywordstyle=\color{codekeyword}\bfseries,
  comment=[l]{--},
  commentstyle=\color{codecomment}\itshape,
  string=[b]",
  stringstyle=\color{codestring},
  morekeywords=[2]{Stage 1, Stage 2, Stage 3, Stage 4},
  keywordstyle=[2]\color{codestage}\bfseries,
  sensitive=true
}

\begin{figure}[t!]
\begin{lstlisting}[
  language=LuauCustom,    
  basicstyle=\footnotesize\ttfamily,
  breaklines=true,
  numbers=left,
  xleftmargin=2em,
  frame=single,
  framesep=4pt
]
local function makeDroneFunctional(root, parts, config)
    local body, blades, legs, button = classifyParts(parts)

    -- Stage 1: Welding -- assemble rigid body
    weldPartsTogether(root, body, legs, button)

    -- Stage 2: Rigging -- hinge constraints for propeller axes
    for _, blade in ipairs(blades) do
        attachHinge(body, blade)
    end

    -- Stage 3: Dynamic force control -- physics-based thrust
    local flightData = setupPhysicsFlight(root, body, blades)
    RunService.Heartbeat:Connect(function(dt)
        updateThrust(flightData, dt)
        updateBladeTilt(blades, flightData.moveDirection)
    end)

    -- Stage 4: Interaction -- bind player input
    bindClickDetector(parts, function() toggleFlight(flightData) end)
    bindMoveEvent(root, flightData)
end
\end{lstlisting}
\vspace{-10 pt}
\caption{Skeleton of a drone behavior script, illustrating the four-stage pipeline. The example targets the Lua scripting API of a commercial gaming platform; the four-stage structure transfers to other engines with comparable rigid-body and event APIs.}
\vspace{-10pt}
\label{fig:drone-skeleton}
\end{figure}

\section{Behavior Script pipeline}

Using a generated drone as an example, we show the skeleton of a behavior script in \reffig{drone-skeleton}, following four core stages:
\begin{enumerate}
    \item \textbf{Welding}: Assembles modular parts into a cohesive rigid body.
    \item \textbf{Rigging}: Configures hinge constraints to define pivot points, e.g., propeller axes.
    \item \textbf{Dynamic Force Control}: Computes real-time physical thrust based on angular velocity and lift formulas.
    \item \textbf{Interaction}: Synchronizes user commands to translate player input into physical motion. 
\end{enumerate}
The controllable part structure allows developers to easily inject category-specific behavior functions.

\section{Additional results}

\reffig{result_supp} presents a broader gallery of generated assets, showcasing the diversity across various object categories and complex part schemas. Furthermore, we include a video that highlights the practical utility of our generated multi-part meshes. 

\begin{figure*}[!htbp]
    \centering
    \vspace{-10 pt}
    \includegraphics[width=\linewidth]{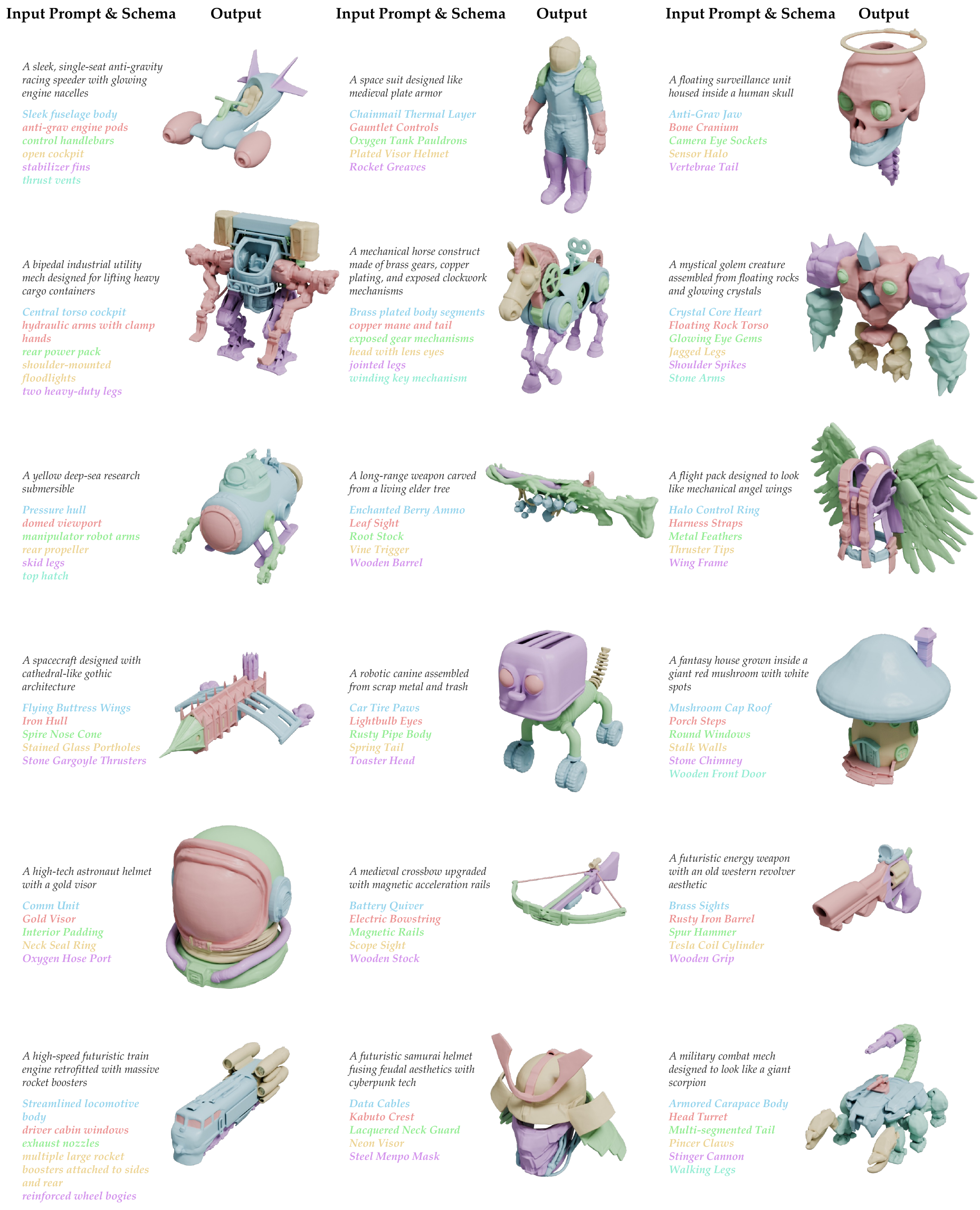}
    \vspace{-25 pt}
    \caption{\textbf{Additional Results.} We present more results generated by our full pipeline.}
    \lblfig{result_supp}
\end{figure*}

\end{document}